%% file: main.tex
\documentclass[
    10pt,
    twocolumn,
    ]{article}

\usepackage{titling}
\usepackage[utf8]{inputenc}

\usepackage[
    colorlinks=true,
    linkcolor=black,
    citecolor=black,
    filecolor=black,
    urlcolor=black,
    plainpages=false,
    pdfpagelabels,
    breaklinks=true,
    ]{hyperref}

\usepackage[
    natbib=true,
    backend=bibtex,
    style=nature,
    citestyle=numeric-comp,
    sorting=none,
    giveninits=true,
    uniquename=init
]{biblatex}
\renewbibmacro*{name:andothers}{
    \ifboolexpr{
    test {\ifnumequal{\value{listcount}}{\value{liststop}}}
    and
    test \ifmorenames
  }
    {\ifnumgreater{\value{liststop}}{1}
       {\finalandcomma}
       {}%
     \andothersdelim\bibstring{andothers}}
    {}}
\usepackage{amsthm,amsmath}
\usepackage{amsfonts, amssymb, amscd}
\usepackage{bookmark}
\bookmarksetup{depth=4}
\usepackage{bbm}
\usepackage[font=footnotesize,labelfont=bf]{caption}
\usepackage{cuted}
\usepackage{enumitem}
\usepackage[capitalize, nameinlink]{cleveref}
\newcommand{\crefp}[1]{\cref{#1}} %
\newcommand{\crefpPlTwo}[2]{\cref{#1,#2}} %
\crefname{equation}{Eqn.}{Eqns.}
\Crefname{equation}{Eqn.}{Eqns.}
\crefname{methods}{Methods}{Methods}
\crefname{supplementary}{Supplementary Information}{Supplementary Information}
\crefname{supplementarytable}{Supplementary Information, Table}{Supplementary Information, Tables}

\usepackage{geometry}
\usepackage{setspace}
\usepackage{glossaries-extra}
\setabbreviationstyle[acronym]{long-short}
\glssetcategoryattribute{acronym}{nohyperfirst}{true}

\usepackage{graphicx}
\usepackage{listings}                            %
\usepackage{mathtools}
\usepackage{multicol}
\usepackage{multirow}
\usepackage{placeins}  %
\usepackage{relsize}
\usepackage{sidecap}
\usepackage[binary-units]{siunitx}
\usepackage{threeparttable}
\usepackage{tikz}
\usepackage{lineno}

\usepackage{xcolor}
\usepackage{xspace}

\input{glossaries.tex}

\newcommand*\subtxt[1]{_{\textnormal{#1}}}
\DeclareRobustCommand\_{\ifmmode\expandafter\subtxt\else\textunderscore\fi}

\newcommand{\br}[1]{{(#1)}}
\newcommand{\bss}{\mbox{BrainScaleS-1}\xspace}
\newcommand{\dls}{\mbox{BrainScaleS-2}\xspace}
\newcommand{\dlsNobox}{BrainScaleS-2}

\newcommand{\gl}{\gL}
\newcommand{\gL}{g_\ell}
\newcommand{\T}{T}
\newcommand{\taum}{\tau\_{m}}
\newcommand{\tauref}{\tau\_{ref}}
\newcommand{\taus}{\tau\_{s}}

\newcommand{\vect}[1]{\mathbf{#1}}
\newcommand{\Vleak}{E_\ell}
\newcommand{\Vth}{\vartheta}

\newcommand{\methods}{\hyperref[sec:methods]{Methods}}

\newcommand{\frach}[2]{\frac{#1}{#2}}
\newcommand{\fracl}[2]{{{#1}/{#2}}}

\definecolor{mpl_c0}{rgb}{0.12156862745098039, 0.4666666666666667, 0.7058823529411765}
\definecolor{mpl_c1}{rgb}{1.0, 0.4980392156862745, 0.054901960784313725}
\definecolor{mpl_c2}{rgb}{0.17254901960784313, 0.6274509803921569, 0.17254901960784313}
\definecolor{mpl_c3}{rgb}{0.8392156862745098, 0.15294117647058825, 0.1568627450980392}
\colorlet{correctLabel}{mpl_c3}

\definecolor{lightgrey}{gray}{0.9}
\newcommand{\inlinecode}[2]{\colorbox{lightgrey}{\lstinline[language=#1]$#2$}}

\newcommand{\tikzcircle}[2][red,fill=red]{\tikz[baseline=-0.5ex]\draw[#1,radius=#2] (0,0) circle ;}%
\newcommand{\tikzsquare}[1][red,fill=red]{\tikz[baseline=-0.0ex]\draw[#1] (0,0) -- (0.2cm, 0) -- (0.2cm,0.2cm) -- (0,0.2cm) -- cycle;}%
\newcommand{\tikztriangle}[1][red,fill=red]{\tikz[baseline=-0.0ex]\draw[#1] (0,0) -- (0.2cm, 0) -- (0.1cm,0.2cm) -- cycle;}%
\newcommand{\tikzdiamond}[1][red,fill=red]{\tikz[baseline=-0.4ex]\draw[#1] (0,0) -- (0.08cm, 0.14cm) -- (0.16cm,0.0cm) -- (0.08cm,-0.14cm) -- cycle;}%


\mathchardef\ordinarycolon\mathcode`\:
    \mathcode`\:=\string"8000
    \begingroup \catcode`\:=\active
    \gdef:{\mathrel{\mathop\ordinarycolon}}
    \endgroup

\date{}

\begin{document}

\title{\textbf{Fast and energy-efficient neuromorphic deep learning\\with first-spike times}}

\newcommand{\affilKIP}{\textsuperscript{1}}
\newcommand{\affilDP}{\textsuperscript{2}}
\newcommand{\affilSiemens}{\textsuperscript{3}}
\newcommand{\correspondingAuthor}{$^{\mathparagraph}$}

\author{
    \normalsize{
        J. Göltz$^{*,}$\correspondingAuthor$^{,}$\affilKIP$^{,}$\affilDP,
        L. Kriener$^{*,}$\correspondingAuthor$^{,}$\affilDP,}\\
    \normalsize{
        A. Baumbach\affilKIP,
        S. Billaudelle\affilKIP,
        O. Breitwieser\affilKIP,
        B. Cramer\affilKIP,
        D. Dold\affilKIP$^{,}$\affilSiemens,
        A. F. Kungl\affilKIP,}
    \\\normalsize{
        W. Senn\affilDP,
        J. Schemmel\affilKIP,
        K. Meier$^{\dagger,}$\affilKIP,
        M. A. Petrovici\correspondingAuthor$^{,}$\affilDP$^{,}$\affilKIP}\\
        \vspace{5pt}
    \footnotesize{
        $^{*}$ Shared first authorship
        \hspace{3em}
        $\dagger$ Deceased
        }\\[-8pt]
    \footnotesize{
        \correspondingAuthor Corresponding authors (julian.goeltz@kip.uni-heidelberg.de, \{laura.kriener,mihai.petrovici\}@unibe.ch)
        }\\[-3pt]
    \footnotesize{
        \affilKIP
        Kirchhoff-Institute for Physics, Heidelberg University, 69120 Heidelberg, Germany.
        }\\[-3pt]
    \footnotesize{
        \affilDP
        Department of Physiology, University of Bern, 3012 Bern, Switzerland.
        }\\[-3pt]
    \footnotesize{
        \affilSiemens
        Siemens AI lab, Siemens AG Technology, 80331 Munich, Germany.
        }\vspace{-8pt}
}

\maketitle
\input{content/abstract.tex}

\section*{Introduction}
\addcontentsline{toc}{section}{Introduction}
\input{content/introduction.tex}

\input{content/outline.tex}

\section*{Results}\label{sec:model}
\addcontentsline{toc}{section}{Results}

\input{content/theory.tex}

\input{content/fig_training.tex}

\paragraph{Simulations}
\label{sec:simulation-emulation}
\input{content/simulations}

\label{sec:learning}
\input{content/sampleSet.tex}

\input{content/mnist.tex}

\paragraph{Fast neuromorphic classification}\label{sec:hw}
\input{content/hardware.tex}

\input{content/bss2.tex}

\paragraph{Robustness of time-to-first-spike learning}\label{sec:stability}
\input{content/stability.tex}

\section*{Discussion}\label{sec:discussion}
\addcontentsline{toc}{section}{Discussion}
\input{content/discussion.tex}

\FloatBarrier
\newpage

\appendix

\section*{Methods}\label{sec:methods}
\setcounter{table}{0}
\setcounter{figure}{0}
\renewcommand{\thefigure}{\Alph{figure}}
\renewcommand{\thetable}{\Alph{table}}
\addcontentsline{toc}{section}{Methods}

    \input{content/methods_calculations.tex}

    \paragraph{\dlsNobox}
    \input{content/methods_emul.tex}
    
    \input{content/methods_simul.tex}

\subsection*{Data availability}
Data available on request from the authors.
\subsection*{Code availability}
Code of the Yin-Yang data set \cite{kriener2021yinyang} available at \url{https://github.com/lkriener/yin_yang_data_set},
other code available on request from the authors.

\FloatBarrier
\printbibliography
\addcontentsline{toc}{section}{References}

\section*{Acknowledgment}\label{sec:ack}
\input{content/ack.tex}

\section*{Author contributions}\label{sec:authcont}
\input{content/authcont.tex}

\section*{Competing Interests statement}
The authors declare no competing interests.

\FloatBarrier
\clearpage
\section*{Supplementary Information}
\glsresetall
\addcontentsline{toc}{section}{Supplementary Information}
\addtocontents{toc}{\protect\setcounter{tocdepth}{0}}
\renewcommand{\thesubsection}{SI.\Alph{subsection}}
\renewcommand{\thefigure}{\thesubsection\arabic{figure}}
\renewcommand{\theequation}{\thesubsection\arabic{equation}}
\renewcommand{\thetable}{\thesubsection\arabic{table}}
\setcounter{figure}{0}
\setcounter{equation}{0}
\setcounter{subsection}{0}
\setcounter{table}{0}
\setcounter{page}{1}

\begin{refsection}

    \FloatBarrier \setcounter{equation}{0} \setcounter{figure}{0} \setcounter{table}{0}
    \subsection{Learning with time-to-first-spike (TTFS) coding on \bss}\label[supplementary]{sec:appBSS1}
    \input{content/SI_bss1.tex}

    \FloatBarrier \setcounter{equation}{0} \setcounter{figure}{0} \setcounter{table}{0}
    \subsection{Additional experiments}\label[supplementary]{sec:additional_experiments}
    \input{content/SI_additional_results.tex}

    \FloatBarrier \setcounter{equation}{0} \setcounter{figure}{0} \setcounter{table}{0}
    \subsection{Robustness to post-training variations}\label[supplementary]{sec:SI_robustnessPostTrain}
    \input{content/SI_post_train_robustness.tex}

    \FloatBarrier \setcounter{equation}{0} \setcounter{figure}{0} \setcounter{table}{0}
    \subsection{Simplification of the learning rule}\label[supplementary]{sec:SI_simpleLR}
    \input{content/SI_LRsimplified}

    \FloatBarrier \setcounter{equation}{0} \setcounter{figure}{0} \setcounter{table}{0}
    \subsection{Power consumption and execution time measurements}\label[supplementary]{sec:SI_energy}
    \input{content/SI_energy}

    \FloatBarrier \setcounter{equation}{0} \setcounter{figure}{0} \setcounter{table}{0}
    \subsection{Extended literature comparison}
    In Table \ref{table:appendix_fullLiterature} we provide a more comprehensive overview of neuromorphic classifiers, including references which lack energy and/or time measurements .
    \input{content/SI_table}

\printbibliography[heading=subbibliography]
\end{refsection}

\end{document}

%% file: glossaries.tex
\newacronym{ppu}{PPU}{plasticity processing unit}
\newacronym{padi}{PADI}{parallel driver interface}

\newacronym{sram}{SRAM}{static random-access memory}
\newacronym{adc}{ADC}{analog-to-digital converter}
\newacronym{cadc}{CADC}{column-parallel analog-to-digital converter}
\newacronym{dac}{DAC}{digital-to-analog converter}
\newacronym{dut}{DUT}{design under testing}
\newacronym{asic}{ASIC}{application-specific integrated circuit}
\newacronym{fpga}{FPGA}{field-programmable gate array}
\newacronym{vlsi}{VLSI}{very-large-scale integration}
\newacronym{simd}{SIMD}{single instruction, multiple data}
\newacronym{ocp}{OCP}{Open Core Protocol}
\newacronym{soc}{SoC}{system on a chip}
\newacronym{sta}{STA}{static timing analysis}
\newacronym{gals}{GALS}{globally asynchronous locally synchronous}
\newacronym{dpi}{DPI}{SystemVerilog direct programming interface}
\newacronym{pll}{PLL}{phase-locked loop}
\newacronym{fifo}{FIFO}{First-In-First-Out Buffer}

\newacronym{psc}{PSC}{postsynaptic current}
\newacronym{psp}{PSP}{postsynaptic potential}
\newacronym{stp}{STP}{short-term plasticity}
\newacronym{stdp}{STDP}{spike-timing-dependent plasticity}
\newacronym{rstdp}{R-STDP}{reward-modulated spike-timing-dependent plasticity}
\newacronym{adex}{AdEx}{adaptive exponential leaky integrate-and-fire}

\newacronym{mc}{MC}{Monte Carlo}

\newacronym{cuba}{CuBa}{current-based}
\newacronym{coba}{CoBa}{conductance-based}
\newacronym{nlif}{nLIF}{non-leaky integrate-and-fire}
\newacronym{lif}{LIF}{leaky integrate-and-fire}
\newacronym{rl}{RL}{reinforcement learning}
\newacronym{ttfs}{TTFS}{time-to-first-spike}
\newacronym{ann}{ANN}{artificial neural network}
\newacronym{snn}{SNN}{spiking neural network}

%% file: content/abstract.tex
\addcontentsline{toc}{section}{Abstract}
\begin{abstract}

    \emph{
        For a biological agent operating under environmental pressure, energy consumption and reaction times are of critical importance.
        Similarly, engineered systems are optimized for short time-to-solution and low energy-to-solution characteristics.
        At the level of neuronal implementation, this implies achieving the desired results with as few and as early spikes as possible.
        With time-to-first-spike coding both of these goals are inherently emerging features of learning.
        Here, we describe a rigorous derivation of a learning rule for such first-spike times in networks of leaky integrate-and-fire neurons, relying solely on input and output spike times, and show how this mechanism can implement error backpropagation in hierarchical spiking networks.
        Furthermore, we emulate our framework on the BrainScaleS-2 neuromorphic system and demonstrate its capability of harnessing the system's speed and energy characteristics.
        Finally, we examine how our approach generalizes to other neuromorphic platforms by studying how its performance is affected by typical distortive effects induced by neuromorphic substrates.
    }
\end{abstract}

%% file: content/introduction.tex
In recent years, the machine learning landscape has been dominated by deep learning methods.
Among the benchmark problems they managed to crack, some were thought to still remain elusive for a long time \citep{krizhevsky2012imagenet, silver2017mastering, brown2020language}.
It is thus not exaggerated to say that deep learning dominates our understanding  of ``artificial intelligence'' \citep{brooks2012brain, ng2016artificial, hassabis2017neuroscience, sejnowski2018deep, richards2019deep}.

Compared to abstract neural networks used in deep learning, their more biological archetypes --- spiking neural networks --- still lag behind in performance and scalability \citep{pfeiffer2018deep}.
Reasons for this difference in success are numerous;
for instance, unlike abstract neurons, even an individual biological neuron represents a complex system, with finite response times, membrane dynamics and spike-based communication~\citep{gerstner2001different, izhikevich2004model}, making it more challenging to find reliable coding and computation paradigms \citep{gerstner1998spiking, maass2016searching, davies2019benchmarks}.
Furthermore, one of the major driving forces behind the success of deep learning, the backpropagation of errors algorithm \citep{linnainmaa1970representation,werbos1982applications,Rumelhart1986}, remained incompatible with spiking neural networks until only very recently \citep{tavanaei2018deep, neftci2019surrogate}.

Despite these challenges, spiking neural networks promise to hold some important advantages.
The time information inherent to spikes allows a coding scheme for spike-based communication that utilizes both spatial and temporal dimensions \citep{gutig2006tempotron}, unlike  spike-count-based approaches \citep{cao2015spiking, diehl2016conversion, schmitt2017neuromorphic, wu2019spikecount}, where the information of spike times is at least partially diluted due to temporal or population averaging.
Owing to the inherent parallelism of all biological, as well as many biologically-inspired, spiking neuromorphic systems \cite{thakur2018large}, this promises fast, sparse and energy-efficient information processing, and provides a blueprint for computing architectures that could one day rival the efficiency of the brain itself \citep{mead1990neuromorphic, roy2019towards, pfeiffer2018deep, thakur2018large}.
This makes spiking neural networks implemented on specialised neuromorphic devices potentially more powerful --- at least in principle --- than the ``conventional'', simple machine learning models currently used on von-Neumann machines, even though this potential still remains mostly unexploited \citep{pfeiffer2018deep}.

Many attempts have been made to reconcile spiking neural networks with their abstract counterparts in terms of functionality, e.g., featuring spike-based  inference models \citep{petrovici2013stochastic, neftci2014event, petrovici2016stochastic, neftci2016stochastic, leng2018spiking, kungl2019accelerated, dold2019stochasticity, jordan2019deterministic,hunsberger2016training} and deep models trained on target spike times by shallow learning rules \citep{kheradpisheh2018stdp, illing2019biologically} or using spike-compatible versions of the error backpropagation algorithm \citep{bohte2000spikeprop, zenke2018superspike, huh2018gradient}.
Especially for tasks operating on static information, a particularly elegant way of utilizing the temporal aspect of exact spike times is the \gls{ttfs} coding scheme \citep{thorpe2001spike}.
Here, a neuron encodes its real-valued response to a stimulus
as the time elapsed before its first spike in reaction to that stimulus.
Such single-spike coding enables fast information processing by explicitly encouraging the emission of as few spikes as early as possible, which meets physiological constraints and reaction times observed in humans and animals \citep{thorpe1996speed,thorpe2001spike,johansson2004first,gollisch2008rapid}.
Apart from biological plausibility, such a fast and sparse coding scheme is a natural fit for neuromorphic systems that offer energy-efficient and fast emulation of spiking neural networks \citep{schemmel2010wafer,akopyan2015truenorth,billaudelle2019versatile,davies2018loihi,mayr2019spinnaker,pei2019towards,moradi2017scalable}.

For hierarchical \gls{ttfs} networks, a gradient-descent-based learning rule was proposed in \citep{mostafa2017supervised,kheradpisheh2020s4nn}, using error backpropagation on a continuous function of output spike times.
However, this approach is limited to a neuron model without leak, which is neither biologically plausible, nor compatible with most analog \gls{vlsi} neuron dynamics~\citep{thakur2018large}.
We propose a solution for \gls{lif} neurons with \gls{cuba} synapses --- a widely-used dynamical model of spiking neurons with realistic integration behavior \citep{rauch2003neocortical,gerstner2009good,teeter2018generalized}.
An early version of this work was presented in \citet{goeltz2019mastersthesis}.

For several specific configurations of time constants, we provide analytical expressions for first-spike timing, which, in turn, allow the calculation of exact gradients of any differentiable cost function that depends on these spike times.
In hierarchical networks of \gls{lif} neurons using the \gls{ttfs} coding scheme, this enables exact error backpropagation, allowing us to train such networks as universal classifiers on both continuous and discrete data spaces.

As our algorithm only requires knowledge about afferent and efferent spike times of all neurons, it lends itself to emulation on neuromorphic hardware.
The accelerated, yet power-efficient \dls platform \citep{friedmann2016hybridlearning,billaudelle2019versatile} pairs especially well with the sparseness and low latency already inherent to \gls{ttfs} coding.
We show how an implementation of our algorithm on \dls can obtain similar classification accuracies to software simulations, while
displaying highly competitive time and power characteristics, with a combination  of \SI{48}{\micro\second} and \SI{8.4}{\micro\joule} per classification.

By incorporating information generated on the hardware for updates during training, the algorithm automatically adapts to potential imperfections of neuromorphic circuits, as implicitly demonstrated by our neuromorphic implementation.
In further software simulations, we show that our model deals well with various levels of substrate-induced distortions such as fixed-pattern noise and limited parameter precision and control, thus providing a rigorous algorithmic backbone for a wide range of neuromorphic substrates and applications. 
Such robustness  with respect to  imperfections of the underlying neuronal substrate represents an indispensable property for any network model aiming for biological plausibility and for every application geared towards physical computing systems \citep{prodromakis2010review,esser2015backpropagation,van2018organic,wunderlich2019demonstrating,kungl2019accelerated,dold2019stochasticity,feldmann2019all}.

%% file: content/outline.tex
In the following, we first introduce the \gls{cuba} \gls{lif} model and the \gls{ttfs} coding scheme, before we demonstrate how both inference and training via error backpropagation can be performed analytically with such dynamics.
Finally, the presented model is evaluated both in software simulations and neuromorphic emulations, before studying effects of several types of substrate-induced distortions.

%% file: content/theory.tex
\begin{figure}[t]
    \centering
    \includegraphics[width=0.48\textwidth]{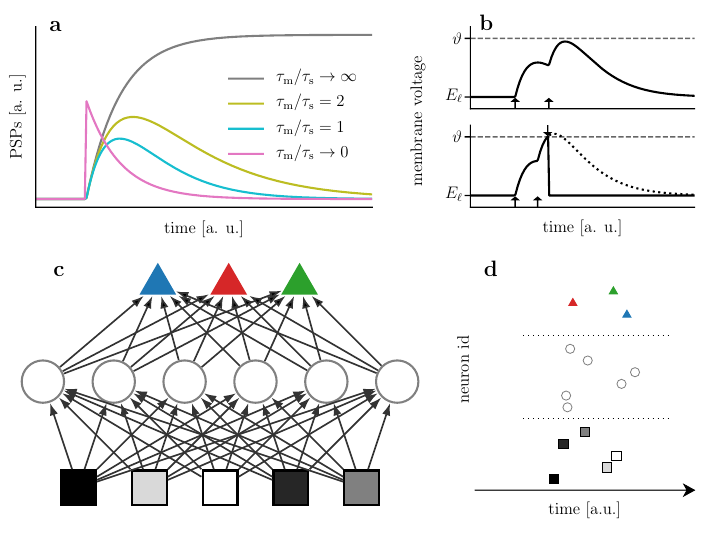}
	\caption{
    	\textbf{Time-to-first-spike coding and learning.}
    	\textbf{Top: single neurons.}
    	\textbf{(a)}
		    \Glsxtrfull{psp} shapes for different ratios of time constants $\taus$ and $\taum$.
	        The finiteness of time constants causes the neuron to gradually forget prior input.
		\textbf{(b)}
		    One key challenge of this finite memory arises when small variations of the synaptic weights result in disappearing/appearing output spikes, which elicits a discontinuity in the function describing output spike timing.
		\textbf{Bottom: application to feedforward hierarchical networks.}
		\textbf{(c)}
    		Network structure.
    		The geometric shape of the neurons represents a notation of their respective types (input {\scriptsize $\square$}, hidden {\normalsize $\circ$}, label {\scriptsize $\triangle$}).
    		The shading of the input neurons is a representation of the corresponding data, such as pixel brightness ({\scriptsize $\blacksquare, \ldots, \textcolor{gray}{\blacksquare}, \ldots, \square$}).
    		The color of the label neurons represents their respective class ($\textcolor{mpl_c0}{\blacktriangle}$, $\textcolor{mpl_c3}{\blacktriangle}$, $\textcolor{mpl_c2}{\blacktriangle}$).
		\textbf{(d)}
            \Glsxtrfull{ttfs} coding exemplified in a raster plot.
            As an example of input encoding, the brightness of an input pixel is encoded in the lateness of a spike.
    		Note that in our framework, \gls{ttfs} coding simultaneously refers to two individual aspects, namely the input-to-spike-time conversion and the determination of the inferred class by the identity of the first label neuron to fire ($\textcolor{correctLabel}{\blacktriangle}$).
    	 In all figures we denote units in square brackets; in particular, we use [a. u.] for arbitrary units, and $[1]$ for dimensionless quantities, and [$\taus$] for times that are measured in multiples of the synaptic time constant $\taus$.
	}	
    \label{fig:setup}
\end{figure}

\paragraph{Leaky integrate-and-fire dynamics}
The dynamics of an \gls{lif} neuron with \gls{cuba} synapses are given by
\begin{equation}
	C_\mathrm{m} \dot{u}(t) = \gl [\Vleak - u(t)] + \sum_i w_i \sum_{t_i} \theta(t-t_i) \exp \left( -\frac{t-t_i}{\taus} \right)\,, \label{eq:cubaLIF}
\end{equation}
with membrane capacitance $C\_m$, leak conductance $\gl$ (from which the membrane time constant $\taum=C\_m/\gl$ follows),
presynaptic weights $w_{i}$ and spike times $t_i$, synaptic time constant $\taus$ and $\theta$ the Heaviside step function.
The first sum runs over all presynaptic neurons while the second sum runs over all spikes for each presynaptic neuron.
The neuron elicits a spike at time $T$ when the presynaptic input pushes the membrane potential above a threshold $\Vth$.
After spiking, a neuron becomes refractory for a time period $\tau_\mathrm{ref}$, which is modeled by clamping its membrane potential to a reset value $\varrho$: $u(t') = \varrho$ for $T \leq t' \leq T+\tau_\mathrm{ref}$.   
For convenience and without loss of generality, we set the leak potential $\Vleak = 0$.
\Cref{eq:cubaLIF} can be solved analytically and yields subthreshold dynamics as described by \cref{eq:appendix_lif}.
The choice of $\taum$ and $\taus$ ultimately influences the shape of a \gls{psp}, starting from a simple exponential ($\taum \ll \taus$), to a difference of exponentials (with an alpha function for the special case of $\taum = \taus$) to a graded step function ($\taum \gg \taus$) (\cref{fig:setup}a).
Note that all of these scenarios are conserved under exchange of $\taus$ and $\taum$, as is apparent from the symmetry of the analytical solution (\cref{eq:appendix_lif}).

The first two cases with finite membrane time constant $\taum$ are markedly different from the last one, which is also known as either the \gls{nlif} or simply integrate-and-fire (IF) model and was used in previous work \citep{mostafa2017supervised}.
In the \gls{nlif} model, input to the membrane is never forgotten until a neuron spikes, as opposed to the \gls{lif} model, where the \gls{psp} reaches a peak after finite time and subsequently decays back to its baseline.
In other words, presynaptic spikes in the \gls{lif} model have a purely local effect in time, unlike in the \gls{nlif} model, where only the onset of a \gls{psp} is localized in time, but the postsynaptic effect remains forever, or until the postsynaptic neuron spikes.
A pair of finite time constants thus assigns much more importance to the time differences between input spikes and introduces discontinuities in the neuronal output that make an analytical treatment more difficult (\cref{fig:setup}b).

\paragraph{First-spike times}
Our spike-timing-based neural code follows an idea first proposed in \citep{mostafa2017supervised}.
Unlike coding in \glspl{ann} and different from spike-count-based codes in \glspl{snn}, this scheme explicitly uses the timing of individual spikes for encoding information.
In \glsxtrfull{ttfs} coding, the presence of a feature in a stimulus is reflected by the timing of a neuron's first spike after the onset of the stimulus
, with earlier spikes representing a more strongly manifested feature.
This has the effect that important information inherently propagates quickly through the network, with potentially only few spikes needed for the network to process an input.
Consequently, this scheme enables  efficient processing of inputs, both in terms of time-to-solution and energy-to-solution (assuming the latter depends, in general on the total number of spikes and the time required for the network to solve, e.g., an input classification problem).

In order to formulate the optimization of a first-spike time~$T$ as a gradient-descent problem, we  derive an analytical expression for $T$.
This is equivalent to finding the time of the first threshold crossing by solving $u(T) = \Vth$ for $T$.
Even though there is no general closed-form solution for this problem, analytical solutions exist for specific cases.
For example, we show that (see \methods)
\begin{align}
    \label{eq:equalTimeEquation}
    \T 
    &= 
    \taus \left\{
        {\frac{b}{a_1} - \mathcal{W}\!\left[
               -\frac{\gL\Vth}{a_1} \exp\left(\frac{b}{a_1}\right)
        \right]}
    \right\}
    && \!\!\text{for } \taum= \taus 
    \; 
    \\\text{and}\nonumber
    \\
    \label{eq:doubleTimeEquation}
    \T 
    &= 
    {2\taus\ln\! \left[
            \frac{2a_1}{a_2 + \sqrt{a_2^2 - 4a_1\gL\Vth}}
        \right]
    }
    && \!\!\text{for }\taum=2\taus
    \; ,
\end{align}
where $\mathcal{W}$ is the Lambert W function and using the shorthand notations $a_n$ and $b$ for sums over the set of causal presynaptic spikes \mbox{$C=\{i \; | \; t_i<T\}$} (see \cref{eq:appendix_aDef,eq:appendix_bDef}).
We note that, when calculating the output spike time for a large number of input neurons, determining $C$ can be computationally intensive (see \methods).
One inherent advantage of physical emulation is the reduction of this calculational burden.

The  above equations are differentiable with respect to synaptic weights and presynaptic spike times.
As will be shown in the following, this directly translates to solving the credit assignment problem and thus allows exact error propagation through networks of spiking neurons.
For easier reading, we focus on one specific case ($\taum=\taus$), but the others can be treated analogously.

\paragraph{Exact error backpropagation with spikes}
Learning in \glspl{snn} requires the ability to relate efferent spiking to both afferent weights and spike times.
For the output spike time of a neuron $k$ with presynaptic partners $i$, the first relationship can be formally described by the derivative of the output spike time with respect to the presynaptic weights (\crefp{eq:appendix_equaltime_notinserted_dw}).
Using certain properties of $\mathcal{W}$, we can find a simple expression that can, additionally, be made to depend on the output spike time $t_k$ itself:
\begin{equation}
    \label{eq:equaltime_dw_reinsert}
    \frac{\partial t_k}{\partial w_{ki}}
     = - \frac{1}{a_1}  \frac{\exp \left( \frac{t_i}{\taus} \right)}{\mathcal{W}(z) + 1} 
     \left(t_k - t_i\right) \, ,
\end{equation}
with $a_1$ and $z$ representing functions of $w_{ki}$ and $t_i$ as defined in \cref{eq:appendix_aDef,eq:appendix_zDef}.
Using the output spike time as additional information optimizes learning in scenarios where the exact neuron parameters are unknown and the real output spike time differs from the one calculated under ideal assumptions, as discussed later.

Second, the capability to relate errors in the output spike time to errors in the input spike times allows us to recursively propagate changes from neurons to their presynaptic partners.
\begin{equation}
    \label{eq:equaltime_dt}
	\frac{\partial t_k}{\partial t_i}
     = - \frac{1}{a_1} 
        \frac{ \exp \left( \frac{t_i}{\taus} \right) }{\mathcal{W}(z) + 1} 
    \frac{w_{ki}}{\taus}
    \left(t_k - t_i - \taus\right)
    \, .
\end{equation}
Together, \cref{eq:equaltime_dw_reinsert,eq:equaltime_dt} effectively and exactly solve the credit assignment problem in appropriately parametrized \gls{lif} networks of arbitrary architecture.

We can now apply the findings above to study learning in a layered network.
\Cref{fig:setup}c shows a schematic of our feedforward networks and their spiking activity.
The input uses the same coding scheme as all other neurons: more prominent features are encoded by earlier spikes.
The output of the network is defined by the identity of the label neuron that spikes first (\cref{fig:setup}d).

We denote by $t^\br{l}_k$ the output spike time of the $k$th neuron in the $l$th layer; for example, in a network with $N$ layers, $t^\br{N}_n$ is the spike time of the $n$th neuron in the label layer.
The weight projecting to the $k$th neuron of layer $l$ from the $i$th neuron of layer $l-1$ is denoted by $w^\br{l}_{ki}$.

To apply the error backpropagation algorithm \citep{linnainmaa1970representation, Rumelhart1986}, we choose a loss function that is differentiable with respect to synaptic weights and spike times.
During learning, the objective is to maximize the temporal difference between the correct and all other label spikes.
The following loss function fulfills the above requirements:
\begin{align}
    L[\vect t^\br{N}, n^*] &= \text{dist} \left( t_{n^*}^{(N)}, t_{n \neq n^*}^{(N)} \right) \nonumber \\
                           &= \log \left[ \sum_n \exp \left( -\frac{t^\br{N}_n - t^\br{N}_{n^*}}{\xi\taus} \right) \right] \; ,
    \label{eq:loss}
\end{align}
where $\vect t^{(N)}$ denotes the vector of label spike times $t^\br{N}_n$, $n^*$ the index of the correct label and $\xi \in \mathbb{R}^+$ is a scaling parameter.
This loss function represents a cross entropy between the true label distribution and the softmax-scaled label spike times produced by the network (see \methods).
Reducing its value therefore increases the temporal difference between the output spike of the correct label neuron and all other label neurons.
Notably, it only depends on the spike time difference and is invariant under absolute time shifts, making it independent of the concrete choice of the experiment start which defines $t=0$.
In case of a non-spiking label neuron we treat its spike time as $t_n^{(N)} = \infty$.
In this case however, the equation \cref{eq:equalTimeEquation} is not defined and neither are its derivatives.
We therefore introduce a simple, local heuristic to encourage spiking behaviour in large portions of the network (see \methods).
In some scenarios, learning can be facilitated by the addition of a spike-time-dependent regularization term (see \methods).

Gradient descent on the loss function \cref{eq:loss} can now be easily performed by repeated application of the chain rule.
Using the exact derivatives \cref{eq:equaltime_dw_reinsert,eq:equaltime_dt}, this yields the synaptic plasticity rule
\begin{align}
    \label{eq:backprop}
    \Delta w^{(l)}_{ki} &\propto 
        - \frac{\partial L[\vect t^{(N)}, n^*]}{\partial w^{(l)}_{ki}} 
        \\
                        &= - \frac{\partial t^{(l)}_k}{\partial w^{(l)}_{ki}}  
                            \underbrace{\frac{\partial L[\vect t^{(N)}, n^*]}{\partial t^{(l)}_{k}}}_{
                                \delta^\br{l}_k}
                        = - \frac{\partial t^{(l)}_k}{\partial w^{(l)}_{ki}} 
                        \mathlarger\sum_j 
                           \frac{\partial t^{(l+1)}_j}{\partial t^{(l)}_k}
                           \delta^\br{l+1}_j
                           \nonumber
       \, .
\end{align}

A compact formulation for hierarchical networks that highlights the backpropagation of errors can be found in \cref{eq:appendix_compactTopError,eq:appendix_compactErrorBP,eq:appendix_compactWeightUpdate}.
In either form,  only the label layer error and the neuron spike times are required for training, which can either be calculated using \cref{eq:equalTimeEquation} or by simulating (or emulating) the \gls{lif} dynamics (\crefp{eq:cubaLIF}).

    The computational complexity of the synaptic plasticity rule -- a potential limiting factor for on-chip implementations -- can be drastically reduced by appropriate approximations.
    In the \cref{sec:SI_simpleLR} we present early results using such an approach.
    Note that the simplification is only used in \cref{sec:SI_simpleLR} and all other results we report in the following were produced
    using the full analytical equations \cref{eq:equaltime_dw_reinsert,eq:equaltime_dt}.

%% file: content/fig_training.tex
\begin{figure*}[ht]
    \centering
    \includegraphics[width=0.96\textwidth]{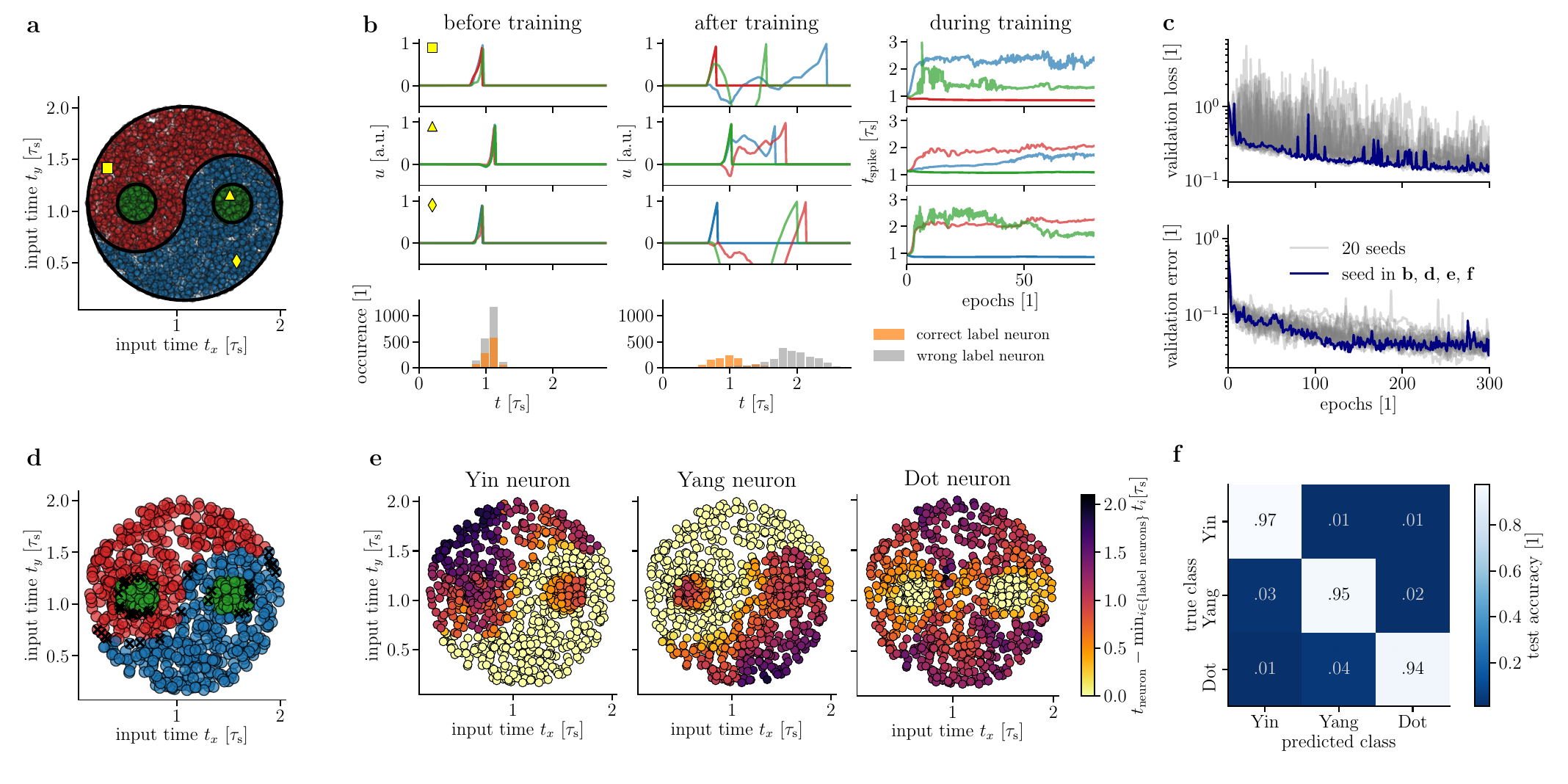}
    \vspace{-0.3cm}
    \caption{
        \textbf{Classification of the Yin-Yang data set.}
        \textbf{(a)}
            Illustration of the Yin-Yang data set.
            The samples are separated into three classes, Yin (\tikzcircle[fill=mpl_c0]{3pt}), Yang (\tikzcircle[fill=mpl_c3]{3pt}) and Dot (\tikzcircle[fill=mpl_c2]{3pt}).
            The yellow symbols (\tikzsquare[fill=yellow], \tikztriangle[fill=yellow], \tikzdiamond[fill=yellow]) mark samples for which the training process is illustrated in (b).
            The input times $t_x$ and $t_y$ correspond to the spike time of the inputs associated with the $x$ and $y$ coordinates of individual samples.
        \textbf{(b)}
            Training mechanism for three exemplary data samples (cf. (a)).
            For the first three rows, the left and middle columns depict voltage dynamics in the label layer before and after training for 300 epochs, respectively.
            The voltage traces of the three label neurons are color-coded according to their corresponding class as in (a).
            Before training, the random initialization of the weights causes the label neurons to show similar voltage traces and almost indistinguishable spike times.
            After training there is a clear separation between the spike time of the correct label neuron and all others, with the correct neuron spiking first.
            The evolution of the label spike times during training is shown in the right column for the first 70 epochs.
            Bottom row: spike histograms over all training samples.
            Our learning algorithm induces a clear separation between the spike times of correct and wrong label neurons.
        \textbf{(c)}
            Training progress (validation loss as given in \cref{eq:loss} and error rate) over 300 epochs for 20 training runs with random initializations (gray).
            The run shown in panels b and d-f is plotted in blue.
        \textbf{(d)}
            Classification result on the test set (1000 samples).
            The color of each sample indicates the class determined by the trained network.
            The wrongly classified samples (marked with black X) all lie very close to the border between classes.
        \textbf{(e)}
            Spike times of the Yin, Yang and Dot neurons for all test samples after training.
            For each sample, spike times were normalized by subtracting the earliest spike time in the label layer.
            Bright yellow denotes zero difference, i.e., the respective label neuron was the first to spike and the sample was assigned to its class.
            The bright yellow areas resemble the shapes of the Yin, Yang and Dot areas, reflecting the high classification accuracy after training.
        \textbf{(f)}
            Confusion matrix for the test set after training.
    }
    \label{fig:trainingChange}
\end{figure*}

%% file: content/simulations.tex
After deriving the learning algorithm in the previous chapter, we show its classification capabilities in software simulations.
In these simulations we demonstrate successful learning and provide a baseline for the hardware emulations that follow.

We use two data sets that emphasize different aspects of interesting real-world scenarios.

As an example for low-dimensional,  ``continuous" data spaces, in which points belonging to different classes can be arbitrarily close together (thus making separation particularly challenging), we chose the Yin-Yang data set \citep{kriener2021yinyang}.
For higher-dimensional, discrete input, we used the MNIST data set \citep{lecun1998gradient} as a small-scale image classification scenario.

%% file: content/sampleSet.tex
The results in this section are based on \cref{eq:equalTimeEquation} for calculating the spike times in the forward pass, and \cref{eq:appendix_compactWeightUpdate} for calculating weight updates;
for details regarding implementation see \methods.
For hyperparameters of the discussed experiments see~\cref{table:sim_params,table:hw_params}.

\emph{Yin-Yang classification task:}
The first data set consists of points in the yin-yang figure (\cref{fig:trainingChange}a).
Each point is defined by a pair of Cartesian
coordinates $(x,y) \in [0,1]^2$.
To build in redundancy and capture the intrinsic symmetry of the yin-yang motive,  the data set is augmented with mirrored coordinates $(1-x,1-y)$ enabling networks of neurons without trainable bias to learn the task \citep{kriener2021yinyang}.
The three classes are labeled as per the respective area they occupy, i.e., Yin, Yang or Dot.
This augmented data set was specifically designed to require latent variables for classification: a shallow non-spiking classifier reaches $(64.3 \pm 0.2)\%$ test accuracy, an \gls{ann} with one hidden layer of size 120 typically around $(98.7 \pm 0.3)\%$.
Due to this large gap, our Yin-Yang data set represents an expressive test of error backpropagation in our hierarchical spiking networks.
At the same time, it can be learned by networks that are compatible in size with the current revision of \dls~\citep{schemmel2020accelerated}.

After translation of the four features to spike times (see \cref{fig:setup} and \methods\ for more details),
they were joined with a bias spike at fixed time, and these five spikes served as input to a network with 120 hidden and 3 label neurons.
We illustrate the training mechanism with voltage traces for three samples belonging to different classes (\cref{fig:trainingChange}b).
The algorithm changes the weights to create a separation in the label spike times (cf. left and middle column) that corresponds to correct classification.
Note that the voltage traces were just recorded for illustration, as only spike times are required for calculating weight updates.
After 300 epochs our networks reached $(95.9\pm0.7)\%$ test accuracy for training with 20 different random seeds (\cref{fig:trainingChange}c).
The classification failed only for samples that were extremely close to the border between two classes (\cref{fig:trainingChange}d).
\Cref{fig:trainingChange}e shows the spike times of the label neurons.
These vary continuously for inputs belonging to other classes, but drop abruptly at the boundary of the area belonging to their own class, which denotes a clear separation -- see, for example, the abrupt change from red (late spike time) to yellow (early spike time) of the Yin-neuron when moving from Yang to Yin (\cref{fig:trainingChange}e, left panel).

%% file: content/mnist.tex
\begin{figure}[ht]
    \centering
    \includegraphics[width=0.48\textwidth]{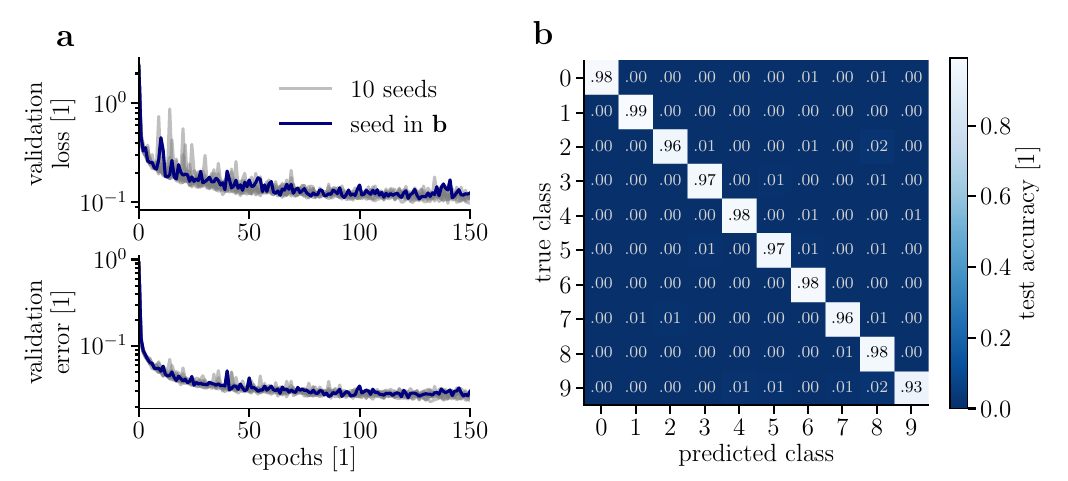}
	\caption{
        \textbf{Classification of the MNIST data set.}
        \textbf{(a)}
            Training progress of a network over 150 epochs for 10 different random initializations.
            The run drawn in blue is the one which produced the results in (b).
        \textbf{(b)}
            Confusion matrix for the test set after training.
	}
    \label{fig:sim_mnist}
\end{figure}

\emph{MNIST classification task:}
To study the scalability of our approach to larger and more high-dimensional data sets, we applied it to the classification of MNIST handwritten digits \citep{lecun1998gradient}.
\Cref{fig:sim_mnist} shows training results for networks with 784-350-10 neurons, where pixel intensities were translated to spike times.
During training, noise was added to the input samples  to aid generalization, but no bias spikes were  used.
As seen in \cref{fig:sim_mnist}a, training converges 
for 10 different initial random seeds, reaching a final test accuracy of
$(97.1\pm0.1)\%$.
Similar results are also achieved for deeper architectures with multiple hidden layers (see \cref{table:additional_results} for additional simulation runs with different network architectures).

For reference, we consider several other results obtained with spiking-time coding.
In \citet{mostafa2017supervised}, a maximum test accuracy of $97.55\%$ using a network with a hidden layer of 800 neurons is reported; note that this work uses non-leaky neurons with effectively infinite membrane memory.
Also for non-leaky neurons, but using an approximative approach for calculating gradients, \citet{kheradpisheh2020s4nn} report $97.4\%$ using 400 hidden neurons.
In \citet{comsa2020temporal}, a maximum test accuracy of $97.96\%$ was achieved using 340 hidden neurons, supported by a regular spike grid and extensive hyperparameter search.

    We note that there also exist trial-averaging and spike-count-based
    approaches that have the benefit of more straight-forward learning rules, but these approaches sacrifice precision, neuronal real-estate or time-to-solution in comparison to frameworks based on the precise timing of single output spikes.

For example, \citet{esser2015backpropagation} report $92.7\%$ using 512 neurons, while \citet{tavanaei2018training} require 1000 hidden neurons to achieve $96.6\%$.

%% file: content/hardware.tex
In our framework, the time to solution is a function of the network depth and the time constants $\taum$ and $\taus$.
Assuming typical biological timescales, most input patterns in the above scenario are classified within several milliseconds.
By leveraging the speedup of neuromorphic systems such as BrainScaleS \citep{schemmel2010wafer,schemmel2020accelerated}, with intrinsic acceleration factors of \numrange{e3}{e4}, the same computation can be achieved within microseconds.
In the following, we present an implementation of our framework on \dls and discuss its performance in conjunction with the achieved classification speed and energy consumption.
For a proof-of-concept implementation on its predecessor \bss, we refer to \cref{sec:appBSS1}.

The advantages of such a neuromorphic implementation come at the cost of reduced control.
Training needs to cope with phenomena such as spike jitter, limited weight range and granularity, as well as neuron parameter variability, among others.
In general, an important aspect of any theory aiming for compatibility with physical substrates, be they biological or artificial, is its robustness to substrate imperfections; our results on \dls implicitly represent a powerful demonstration of this property.
To further substantiate the generalizability of our algorithm to different substrates, we complement our experimental results with a simulation study of various substrate-induced distortive effects.

%% file: content/bss2.tex
\begin{figure*}[ht]
    \centering
    \includegraphics[width=0.98\textwidth]{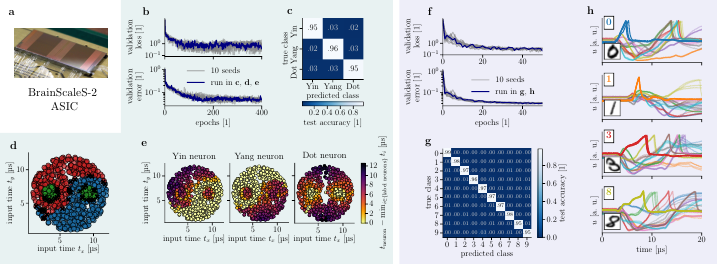}
	\caption{
        \textbf{Classification on the \dls neuromorphic platform.}
        \textbf{(a)}
            Photograph of a \dls chip.
        \textbf{(b-e) Yin-Yang data set}
        \textbf{(b)}
            Training progress over 200 epochs for 11 different random initializations.
            The run drawn in blue also produced the results shown in panel (b-d).
        \textbf{(c)}
            Confusion matrix for the test set after training.
        \textbf{(d)}
            Classification result on the test set.
            For each input sample the color indicates the class determined by the trained network.
            Wrong classifications are marked with a black X.
            The wrongly classified samples all lie very close to the border between two classes.
        \textbf{(e)}
                Separation of label spike times (cf. \cref{fig:trainingChange}e).
                For each of the label neurons, bright yellow dots represent data samples for which it was the first to spike, thereby assigning them its class.
                Similarly to the software simulations, the bright yellow areas align well with the shapes of the Yin, Yang and Dot areas of the data set.
        \textbf{(f-h) MNIST data set}
        \textbf{(f)}
            Evolution of training over 50 epochs for 10 different random initializations.
            The run drawn in blue is the one which produced the results shown in panel (g) and (h).
        \textbf{(g)}
            Confusion matrix for the test set after training.
        \textbf{(h)}
            Exemplary membrane voltage traces on \dls after training.
            Each panel shows color-coded voltage traces of four label neurons for one input that was presented repeatedly to the network (inlays show the input and its correct class).
            Each trace was recorded four times to point out the trial-to-trial variations.
	}
    \label{fig:bss2}
\end{figure*}

\begin{table*}
    \caption{Comparison of pattern recognition models on the MNIST data set emulated on neuromorphic back-ends, sorted by classification speed. 
    For reference, an \gls{ann} running  on GPU is included in the top row.
    Note that we include only references which present measurements for both energy and throughput in addition to accuracy.
    An extended table containing results with partial or estimated measurements can be found in \cref{table:appendix_fullLiterature}.
    }
    \centering
    \begin{threeparttable}
    \resizebox{\textwidth}{!}{%
    \begin{tabular}{llclccccccc}
        \multirow{2}{*}{\textbf{platform}} & \multirow{2}{*}{\textbf{type}}  & \multirow{2}{*}{\textbf{technology}} & \multirow{2}{*}{\textbf{coding}} & \textbf{input} & \textbf{network} & \textbf{data augmentation/} & \textbf{energy per} & \textbf{classifications} & \textbf{test} & \multirow{2}{*}{\textbf{reference}}\\
        &&&& \textbf{resolution} & \textbf{size/structure} & \textbf{regularization} &\textbf{classification} & \textbf{per second}\tnote{1} & \textbf{accuracy} & \\[0.3em]
        \hline
        &&&&&\\[-0.7em]
        Nvidia Tesla P100 & digital & \SI{14}{\nano\meter} & \gls{ann} & $28\times 28$ & CNN\tnote{2} & dropout & \SI{852}{\micro\joule} & $125\,000$ & \SI{99.2}{\percent} & see \labelcref{subsec:SI_energy_vN} \\[1.0em]
        SpiNNaker & digital & \SI{130}{\nano\meter}& rate & $28\times28$ & 784-600-500-10 & noisy input encoding & \SI{3.3}{\milli\joule} & $91$  & \SI{95.0}{\percent} & \cite{stromatias2015scalable}, 2015\\[0.3em]
        True North & digital & \SI{28}{\nano\meter}& rate & $28\times28$ & CNN & noisy input encoding& \SI{0.27}{\micro\joule} & $1000$ & \SI{92.7}{\percent} & \cite{esser2015backpropagation}, 2015 \\[0.3em]
        True North & digital & \SI{28}{\nano\meter} & rate & $28\times28$ & CNN & noisy input encoding & \SI{108}{\micro\joule} & $1000$ & \SI{99.4}{\percent} & \cite{esser2015backpropagation}, 2015\\[0.3em]
        unnamed (Intel) & digital & \SI{10}{\nano\meter}& temporal & ($28\times28$)\tnote{3} & 236-20 & stochastic spike loss & \SI{17.1}{\micro\joule} & $6250$ & \SI{89.0}{\percent} & \cite{chen20184096}, 2018 \\[0.3em]
        \multirow{2}{*}{\dls} & \multirow{2}{*}{mixed} & \multirow{2}{*}{\SI{65}{\nano\meter}}& \multirow{2}{*}{temporal} & \multirow{2}{*}{$16\times16$} & \multirow{2}{*}{256-246-10} & \multirow{2}{*}{input noise} & \multirow{2}{*}{\SI{8.4}{\micro\joule}} & \multirow{2}{*}{$20\,800$} & \multirow{2}{*}{\SI{96.9}{\percent}} & this work,\\
        &&&&&&&&&&see also \labelcref{subsec:SI_energy_bss}\\[1.0em]
    \end{tabular}
    }
    \\
    \begin{tablenotes}
        \item[1] \footnotesize Note that some of the platforms achieve a high number of classifications per second simply by processing a large number of samples in parallel,\\ while other platforms rely on the sequential (but fast) processing of individual samples.
        \item[2] \footnotesize  Standard architecture given as an example in the PyTorch repository, for details see \cref{subsec:SI_energy_vN}.
        \item[3] \footnotesize The $28\times28$ image is preprocessed using $5\times5$ Gabor-filters and $3\times3$ pooling before being sent into the chip.
    \end{tablenotes}
    \end{threeparttable}
    \label{table:energy_reference}
\end{table*}

\emph{Learning on BrainScaleS-2:}
\dls is a mixed-signal accelerated neuromorphic platform with 512 physical neurons, each being able to receive inputs via 256 configurable synapses.
These neurons can be coupled to form larger logical neurons with a correspondingly increased number of inputs.
At the heart of each neuron is an analog circuit emulating \gls{lif} neuronal dynamics with an acceleration factor of \numrange{e3}{e4} compared to biological timescales.

Due to variations in the manufacturing process, the realized circuits systematically deviate from each other (fixed-pattern noise). 
Although these variations can be reduced by calibrating each circuit \citep{aamir2018lifarray}, considerable differences remain (standard deviation on the order of \SI{5}{\percent} on \dls) and pose a challenge for possible neuromorphic algorithms -- along with other features of physical model systems such as spike time jitter or spike loss \citep{petrovici2014characterization, wunderlich2019demonstrating, kungl2019accelerated, dold2019stochasticity}.

The chip's synaptic arrays were configured to support arbitrary fully-connected networks of up to 256 emulated neurons with a maximum of 256 inputs per neuron.
Each such logical connection was realized via two physical synapses in order to allow transitions between an excitatory and an inhibitory regime.
Synaptic weights on the chip are configurable with 6 bit precision.
More details about our setup can be found in the \methods\ section.

We used an in-the-loop training approach \citep{schmitt2017neuromorphic,kungl2019accelerated,cramer2020training}, where inference runs emulated on the neuromorphic substrate were interleaved with host-based weight update calculations.
For emulating the forward pass, the spike times for each sample in a mini-batch were joined sequentially into one long spike train and then injected into the neuromorphic system via a \gls{fpga}.
The latter was also used to record the spikes emitted by the hidden and label layers.

\Cref{fig:bss2}a-d shows the results of training a spiking network with 120 hidden neurons on \dls on the Yin-Yang data set.
The system quickly learned to discriminate between the presented patterns, with an average test accuracy of $(95.0\pm0.9)\%$.

The hardware emulation performs similarly to the software simulations (\cref{fig:trainingChange}), with the wrong classifications still only happening along the borders of the areas with different labels (\cref{fig:bss2}c).
The remaining difference in performance after training is attributable to the substrate variability (cf. also \cref{fig:bss2}h).
Considering that one of the specific challenges built into the Yin-Yang data set resides in the continuity of its input space and abrupt class switch between bordering areas, this result highlights the robustness of our approach.

To classify the MNIST data set using the \dls system, we emulated and trained a network of size 256-246-10 (\cref{fig:bss2}f-h).
Due to the restrictions imposed by the hardware on the input dimensionality, we used downsampled images of $16\times16$ pixels.
Across multiple initializations, we achieved a test accuracy of $(96.9\pm0.1)\%$; similarly to the Yin-Yang data set, this is only slightly lower than in software simulations of equally sized networks (\cref{table:summary_of_results}).
As shown in \cref{table:summary_of_results}, about one third of the loss in accuracy is due to the downsampling of the data, with the remainder being caused by the variability of the substrate.
The ability of our framework to achieve reliable classification despite such substrate-induced distortions is well-illustrated by post-training membrane dynamics measured on the chip \cref{fig:bss2}h.
In all cases shown here, the correct label neuron spikes before \SI{10}{\micro\second} and is clearly separable from all other label neurons.

Due to its short intrinsic time constants and overall energy efficiency, the \dls system enables very fast and energy-efficient acquisition of classification results.
Classification of the 10\,000 MNIST test samples takes a total of \SI{0.937}{\second}, including data transmission, emulation of dynamics and return of the classification results.
The total time on the \dls chip was \SI{480}{\micro\second}, a detailed breakdown of the execution time is shown in \cref{sec:SI_energy}.
The power consumption of the chip, measured during runtime, including all chip components needed for spike generation and processing (i.e., excluding the host and \gls{fpga}) amounted to \SI{175}{\milli\watt}.
For measurement details and scalability considerations we refer to \cref{sec:SI_energy}.
This results in an average energy consumption of \SI{8.4}{\micro\joule} per classification.
For a comparison to other neuromorphic platforms, we refer to \cref{table:energy_reference}.

    Note that the networks on the other neuromorphic platforms differ in their architectures, coding schemes and training methods, and while we list some of these differences in the table, a direct comparison in terms of individual numbers remains difficult.

This table only includes references in which measurements for both classification rate and energy are reported.
A more comprehensive overview, including studies that lack some of the above measurements, can be found in the \cref{table:appendix_fullLiterature}.

Our current experimental setup leaves room for significant optimization.
For an estimation of possible improvements and their potential effect on classification rate and energy consumption, we refer to \cref{sec:SI_energy} and \citep{cramer2020training}.
With these improvements we expect to increase the classification rate by up to a factor of four while simultaneously decreasing the energy-per-classification value by up to a factor of 3.

\input{content/table_setComparison.tex}

%% file: content/table_setComparison.tex
\begin{table}
    \caption{
        Summary of the presented results.
        Accuracies are given as mean value and standard deviation.
        For comparison, on the Yin-Yang data set a linear classifier achieves \SI[parse-numbers=false]{(64.3 \pm 0.2)}{\percent} test accuracy, while a (non-spiking, not particularly optimized) ANN with 120 hidden neurons achieves \SI[parse-numbers=false]{(98.7 \pm 0.3)}{\percent}.
        As a reference for the MNIST data set we trained a 784-350-10 fully connected ANN which reached an average test accuracy of \SI[parse-numbers = false]{(98.2 \pm 0.1)}{\percent}.
        The results in this table were obtained without extensive hyperparameter tuning.
    }
    \centering
    \begin{threeparttable}
    \resizebox{\columnwidth}{!}{
    \begin{tabular}{l|c|c|c}
        \multirow{2}{*}{\textbf{data set}}  & \textbf{hidden} & \multicolumn{2}{c}{\textbf{accuracy $[\si{\percent}]$}} \\
        & \textbf{neurons} & \textbf{test} & \textbf{train} \\[0.3em]
                \hline
        & & & \\[-0.7em]
        \textbf{Yin-Yang} & & & \\[0.3em]
        in SW & 120 & $95.9 \pm 0.7$ & $96.3 \pm 0.7$ \\
        on HW & 120 & $95.0\pm 0.9$ & $95.3\pm 0.7$ \\[0.3em]
        \textbf{MNIST} & & & \\[0.3em]
        in SW & 350 & $97.1 \pm 0.1$ & $99.6 \pm 0.1$ \\
        in SW ($\taus = 2\taum$) & 350 & $97.2 \pm 0.1$ & $ 99.7 \pm 0.1$ \\[0.3em]
        \textbf{MNIST 16$\times$16} & & & \\[0.3em]
        in SW & 246 & $97.4 \pm 0.2$ & $ 99.2 \pm 0.1$ \\
        on HW & 246 & $96.9 \pm 0.1$ & $98.2 \pm 0.1$ \\ 
    \end{tabular}
    }
    \end{threeparttable}
    \label{table:summary_of_results}
\end{table}

%% file: content/stability.tex
As noted earlier, a learning scheme operating only on spike times combined with our coding  represents a natural fit for neuromorphic hardware, both for requiring commonly accessible observables (i.e., spike times, as opposed to, e.g., membrane potentials or synaptic currents)
and due to its intrinsic efficiency, as it emphasizes few and early spikes.
An important indicator of a model's feasibility for neuromorphic emulation is its robustness towards substrate-induced distortions.
By experimentally demonstrating its capabilities on \dls, we have implicitly provided one substantive data point for our framework.
Here, we present a more comprehensive study of the robustness of our approach.

Most physical neuronal substrates have several forms of variability in common \citep[Chapter~5]{petrovici2016form}.
In both digital and mixed-signal systems, synaptic weights are typically limited in both range and resolution.
Additionally, parameters of analog neuron and synapse circuits exhibit a certain spread.

To study the impact of these effects, we included them in software simulations of our model applied to the Yin-Yang classification task.

In this context, we highlight the importance of a detail mentioned in the derivation of \cref{eq:equaltime_dw_reinsert}.
The output spike time given in  \cref{eq:equalTimeEquation} depends only on neuron parameters, presynaptic spike times and weights, thus its derivatives share the same dependencies (\crefpPlTwo{eq:appendix_equaltime_notinserted_dt}{eq:appendix_equaltime_notinserted_dw}).
With some manipulations, the equation for the actual output spike time can be inserted (\crefpPlTwo{eq:appendix_equaltime_reinserted_dt}{eq:appendix_equaltime_reinserted_dw}), producing a version of the learning rule that directly depends on the output spike time itself.
This version thus allows the incorporation of additional information gained in the forward pass and is therefore expected to be significantly more stable, which is confirmed below.

\begin{figure}[ht]
    \centering
    \includegraphics[width=0.48\textwidth]{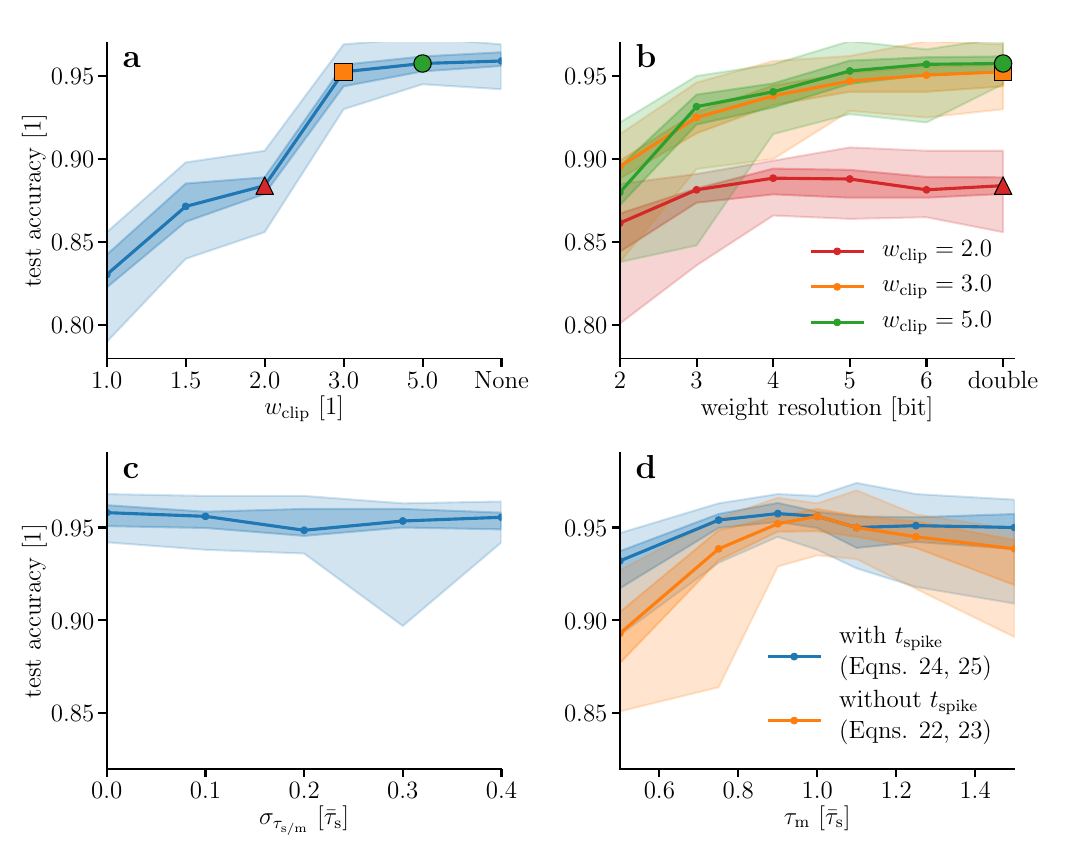}
	\caption{
        \textbf{Effects of substrate imperfections.}
        Modeled constraints were added artificially into simulated networks.
        All panels show median, quartiles, minimum, and maximum of the final test accuracy on the Yin-Yang data set for 20 different initializations.
        \textbf{(a)}
            Limited weight range.
            The weights were clipped to the range $[-w_\text{clip}, w_\text{clip}]$ during training and evaluation.
            The triangle, square and circle mark the clip values that are used in panel (b).
        \textbf{(b)}
            Limited weight resolution.
            For the three weight ranges marked in (a) the weight resolution was reduced from a double precision float value down to 2 bits.
            Here, $n$-bit precision denotes a setup where the interval $[-w_\text{clip}, w_\text{clip}]$ is discretized into $2 \cdot 2^n-1$ samples ($n$ weight bits plus sign).
        \textbf{(c)}
            Time constants with fixed-pattern noise.
            For these simulations each neuron received a random $\tau_\text{s}$ and $\tau_\text{m}$ independently drawn from the distribution $N(\bar\tau_s, \sigma_{\tau_\text{s/m}})$.
            This means that the ratio of time constants was essentially never the one assumed by the learning rule.
        \textbf{(d)}
            Systematic shift between time constants.
            Here $\tau_\text{s}$ was drawn from $N(\bar\tau_\text{s}, \sigma_{\tau_\text{s/m}})$ while $\tau_\text{m}$ was drawn from $N(\bar\tau_\text{m}, \sigma_{\tau_\text{s/m}})$ for each neuron for varying mean $\bar\tau_m$ and fixed $\sigma_{\tau_\text{s/m}}=0.1\bar\tau_\text{s}$.
            The orange curve illustrates a training where the backward pass performs ``naive" gradient descent, without using explicit information about output spike times.
            The blue curve, as all other panels, has the output spike time as an observable.
	}
    \label{fig:stability}
\end{figure}

Using dimensionless weight units (scaled by the inverse threshold), we observe that an upper weight limit of approximately $3$ is sufficient for achieving peak performance (\Cref{fig:stability}a).
This weight value is equivalent to a \gls{psp} that covers the distance between leak potential and firing threshold.

    If this is not achievable within the typical parametrization range of a neuromorphic chip, the effective maximum weight to the hidden layer can be increased by multiplexing each input into the network (cf. \methods).

In the experiments with limited weight resolution (both in software and on hardware), a floating-point-precision ``shadow" copy of synaptic weights was kept in memory.
The forward and backward pass used discretized weight values, while the calculated weight updates were applied to the shadow weights \citep{hubara2017quantized}.
Our model shows approximately constant performance for weight resolutions down to 5 bit, followed by gradual degradation below (\Cref{fig:stability}b).

Interestingly, adding variability to the synapse and membrane time constants has no discernible effects (\Cref{fig:stability}c).
This is a direct consequence of having used the true output spike times for the learning rule in the backward pass.
A comparison to ``naive" gradient descent without this information is shown in (\Cref{fig:stability}d).
These simulations show that the algorithm can be expected to adequately cope with a large amount of fixed-pattern noise on the time constants if the mean of the distributions for $\taum$ and $\taus$ match reasonably well with the values assumed by the learning rule (up to 10-20\% difference).

    Additionally, in \cref{sec:SI_robustnessPostTrain} we investigate trained networks regarding their robustness to adverse effects that appear only after training, such as temperature-induced parameter variations or inactivation of neurons.
    Our simulations show that trained networks can cope with such effects, suggesting that our training algorithm develops network structures robust even to distortions not present during training.

Finally, we note that all of the effects addressed above also have biological correlates.
While not directly reflecting the variability of biological neurons and synapses, our simulations do suggest that biological variability does not present a fundamental obstacle to our form of \gls{ttfs} computation.

%% file: content/discussion.tex
We have proposed a model of first-spike-time learning that builds on a rigorous analysis of neuro-synaptic dynamics with finite time constants and provides exact learning rules for optimizing first-spike times.
The resulting form of synaptic plasticity operates on pre- and postsynaptic spike times and effectively solves the credit assignment problem in spiking networks; for the specific case of hierarchical feedforward topologies, it yields a spike-based form of error backpropagation.
    In this manuscript, we have applied this algorithm to networks with one and two hidden layers.
    Given the reported results, we are confident that our approach scales to even larger and deeper networks.

While \gls{ttfs} coding is an exceptionally appealing paradigm for reasons of speed and efficiency, our approach is not restricted to this particular coding scheme.
Our learning rules enable a rigorous manipulation of spike times and can be used for a variety of loss functions that target other relationships between spike timings.
The time-to-first-spike scenario studied here merely represents the simplest, yet arguably also the fastest and most efficient paradigm for spike-based classification of static patterns.
Additionally, our derived theory is applicable to more complex, e.g., recurrent, network structures and multi-spike coding schemes which are needed for processing  temporal data streams.

First-spike coding schemes are particularly relevant in the context of biology, where decisions often have to be taken under pressure of time.
The action to be taken in response to a stimulus can be considerably sped up by encoding it in first-spike times.
In turn, such fast decision making on the order of $\sim\!\!\SI{100}{ms}$ \citep{thorpe1996speed,thorpe2001spike} will have a particularly sensitive dependence on exact spike times and thus require a corresponding precision of parameters.

At first glance, demands for precision appear at odds with the imperfect, variable nature of microscopic physical substrates, both biological and artificial.
    We met this challenge by incorporating output spike times directly into the backward pass.
    With this, the theoretical requirement of exact ratios of membrane to synaptic time constants is significantly softened, which greatly extends the applicability of our framework to a wide range of substrates, including, in particular, \dls.

By requiring only spike times, the proposed learning framework has minimal demands for neuromorphic hardware and becomes inherently robust towards substrate-induced distortions.
This further enhances its suitability for a wide range of neuromorphic platforms.

Bolstered by the design characteristics of the \dls system, our implementation achieves a time-to-classification of about \SI{10}{\micro\second} after receiving the first spike.
Including relaxation between patterns and communication, the complete MNIST test set with 10\,000 samples is classified in less than \SI{1}{\second} with an energy consumption of about \SI{8.4}{\micro\joule} per classification, which compares favorably with other neuromorphic solutions for pattern classification.
The time characteristics of this implementation do not deteriorate for increased layer sizes because neurons communicate asynchronously and their dynamics are emulated independently.
For the current incarnation of \dls, an increase in spiking activity only has a negligible effect on power consumption.
Furthermore, for larger numbers of neurons we would expect only a weak increase of the power drain.

We also stress that, in contrast to, e.g., GPUs, our system was used to process input data sequentially.
Our reported classification speed is thus a direct consequence of our coding scheme combined with the system's accelerated dynamics.
Further increasing the throughput by parallelization (simultaneously using multiple chips) is straightforward and would not affect the required energy per classification.

    Due to the complexity of our exact gradient-based rules, our hardware networks were trained using updates calculated off-chip based on emulated spike times.
    Early, promising simulations using a significantly simplified learning rule, however, suggest the possibility of an on-chip implementation of our framework.
    Furthermore, we note that our learning rules require three components that can all be made available at the locus of the synapse: pre- and post-synaptic spikes, as in classical spike-timing-dependent plasticity, and an error term, which could be propagated by mechanisms such as those proposed in, e.g.,~\citep{Payeur2020Bursts,sacramento2018}.
    This raises the intriguing possibility for our framework to help explain learning in biological substrates as well.

Since, compared to the von-Neumann paradigm, artificial brain-inspired computing is only in its infancy, its range of possible applications still remains an open question.
This is reflected by most state-of-the-art neuromorphic approaches to information processing, which, in order to accommodate a wide range of spike-based computational paradigms, aim for a large degree of flexibility in network topology and parametrization.
Despite the obvious efficiency trade-off of such general-purpose platforms, we have shown that an embedded version of our framework can achieve a powerful combination of performance, speed, efficiency and robustness.
This gives us confidence 
that a more specialized neuromorphic implementation of our model represents a competitive alternative to current solutions based on von-Neumann architectures, especially in edge computing scenarios.

%% file: content/methods_calculations.tex
    \paragraph{Preliminaries}
    In this section we derive the equations from the main manuscript, starting with the learning rule for $\taum\rightarrow\infty$, then $\taum=\taus$, \cref{eq:equalTimeEquation} and finally $\taum=2\taus$, \cref{eq:doubleTimeEquation}.
    The case $\taum \rightarrow \infty$ has already been discussed in \citet{mostafa2017supervised} and was reproduced here for completeness and comparison.
    Due to the symmetry in $\taum$ and $\taus$ of the \gls{psp} (\crefp{eq:lif_alpha}), the $\taum = 2\taus$ case describes the $\taum = \frac{1}{2}\taus$ case as well.

    For each, a solution for the spike time $T$, defined by 
    \begin{equation}
        \label{eq:def_spike_time}
        u(T)=\Vth,
    \end{equation}
    has to be found, given \gls{lif} dynamics
     \begin{align}
            \label{eq:appendix_lif}
            u(t)
            &= \frac{1}{C\_m} \frac{\taum \taus}{\taum-\taus}
            \sum_{\mathrm{spikes\ } t_i} w_{i}  
            \kappa(t - t_i)\,, 
        \\
            \label{eq:appendix_lif2}
            \kappa(t) &= \theta(t) 
                \left[\exp\left({-\frac{t}{\taum}}\right) - \right.
                \left. \exp\left({-\frac{t}{\taus}} \right)\right] \,,
    \end{align}
    with membrane time constant $\taum = C\_{m} / \gl$ and the \gls{psp} kernel $\kappa$ given by a difference of exponentials.
    Here we already assumed our \gls{ttfs} use case in which each neuron only produces one relevant spike and the second sum in \cref{eq:cubaLIF} reduces to a single term.
    
    For convenience, we use the following definitions
    \begin{align}
        \label{eq:appendix_aDef}
        a_n &:= \sum_{i \in C} w_i \exp\left(\frac{t_i}{n\taus}\right)\,, \\
        \label{eq:appendix_bDef}
        b &:= \sum_{i\in C} w_i\frac{t_i}{\taus} \exp\left(\frac{t_i}{\taus}\right)
            \,,
    \end{align}
    with summation over the set of causal presynaptic spikes $C=\{i \; |\;  t_i<T\}$.
    
    In practice, this definition of the causal set $C$ is not a closed-form expression because the output spike time $T$ depends explicitly on $C$.
    However, it can be computed straightforwardly by
    iterating over the ordered sets of input spike times (for $n$ presynaptic spikes there are $n$ sets $\tilde C_i$ each comprising of the $i$ first input spikes).
    For each set $\tilde C_i$ one calculates an output spike time $T_i$ and determines if this happens later than the last input of this set and before the next input (the $i+1$th input spike).
    The earliest such spike $T_i$ is the actual output spike time and the corresponding $\tilde C_i$ is the correct causal set.
    If no such causal set $\tilde C_i$ exists, the neuron did not spike and we assign it the spike time $T=\infty$.

\paragraph
    [\texorpdfstring{\glsname{nlif}}{nLIF} learning rule \texorpdfstring{for $\taum\rightarrow\infty$}{}]
    {\glsname{nlif} learning rule for $\boldsymbol{\taum\rightarrow\infty}$}
    With this choice of $\taum$, the first term in \cref{eq:appendix_lif2} becomes $1$ and we recover the \gls{nlif} case discussed in \citep{mostafa2017supervised}.
    Given the existence of an output spike, in \cref{eq:def_spike_time} the spike time $T$ appears only in one place and simple reordering yields
    \begin{equation}
        \frac\T\taus 
        = 
        \ln\left[\frac{a_1}{a_\infty - \Vth C\_m/\taus}\right],
    \end{equation}
    where we used \cref{eq:appendix_aDef} for $n=1$ and $n=\infty$, the latter being the sum over the weights.

\paragraph
    [Learning rule for \texorpdfstring{$\taum=\taus$}{equal time constants}]
    {Learning rule for $\boldsymbol{\taum=\taus}$}
    According to l'H\^{o}pital's rule, in the limit $\taum\rightarrow\taus$ \cref{eq:appendix_lif} becomes a sum over $\alpha$-functions of the form
    \begin{equation}
        \label{eq:lif_alpha}
        u(t)
        = \frac{1}{C\_m}
        \sum_{i} w_{i} \theta(t-t_{i})
        \cdot
        (t-t_{i})
        \exp\left({-\frac{t-t_{i}}{\taus}}\right).
    \end{equation}
    Using these voltage dynamics for the equation of the spike time \cref{eq:def_spike_time}, together with the definitions \cref{eq:appendix_aDef,eq:appendix_bDef} and $\taum = C_\mathrm{m} / \gl$, we get the equation
    \begin{equation}
        \label{eq:appendix_et_tmp}
        0 =
        \gL\Vth \exp\left(\frac{T}{\taus} \right)
        +\underbrace{b - a_1 \frac{T}{\taus}}_{=:y}.
    \end{equation}
    The variable $y$ is introduced to bring the equation into the form 
    \begin{equation}
        \label{eq:lambertw_def}
        h \exp\left(h\right)=z
    \end{equation}
    which can be solved with the differentiable Lambert W function $h=\mathcal{W}(z)$.
    The goal is now to bring \cref{eq:appendix_et_tmp} into this form, this is achieved by reformulation in terms of $y$ 
    \begin{gather}
        0 = \gL\Vth \exp\left( \frac{b}{a_1} \right) \exp\left( -\frac{y}{a_1}\right) + y
    \\
        \underbrace{\frac{y}{a_1}}_{\textstyle =:h} \exp\left(\frac{y}{a_1}\right) =
            \underbrace{- \frac{\gL\Vth}{a_1} \exp{\left(\frac{b}{a_1}\right)}}_{\textstyle =: z}
        \,.
    \label{eq:appendix_zDef}
    \end{gather}
    With the definition of the Lambert W function the spike time can be written as 
    \begin{equation}
        \label{eq:appendix_equalTimeEquation}
        \frac{\T}{\taus}
        = 
        \frac{b}{a_1} - \mathcal{W}\!\left[ 
                -\frac{\gL\Vth}{a_1} \exp\left(\frac{b}{a_1}\right)
        \right].
    \end{equation}

    \begin{figure}[ht]
        \centering
        \includegraphics[width=0.48\textwidth]{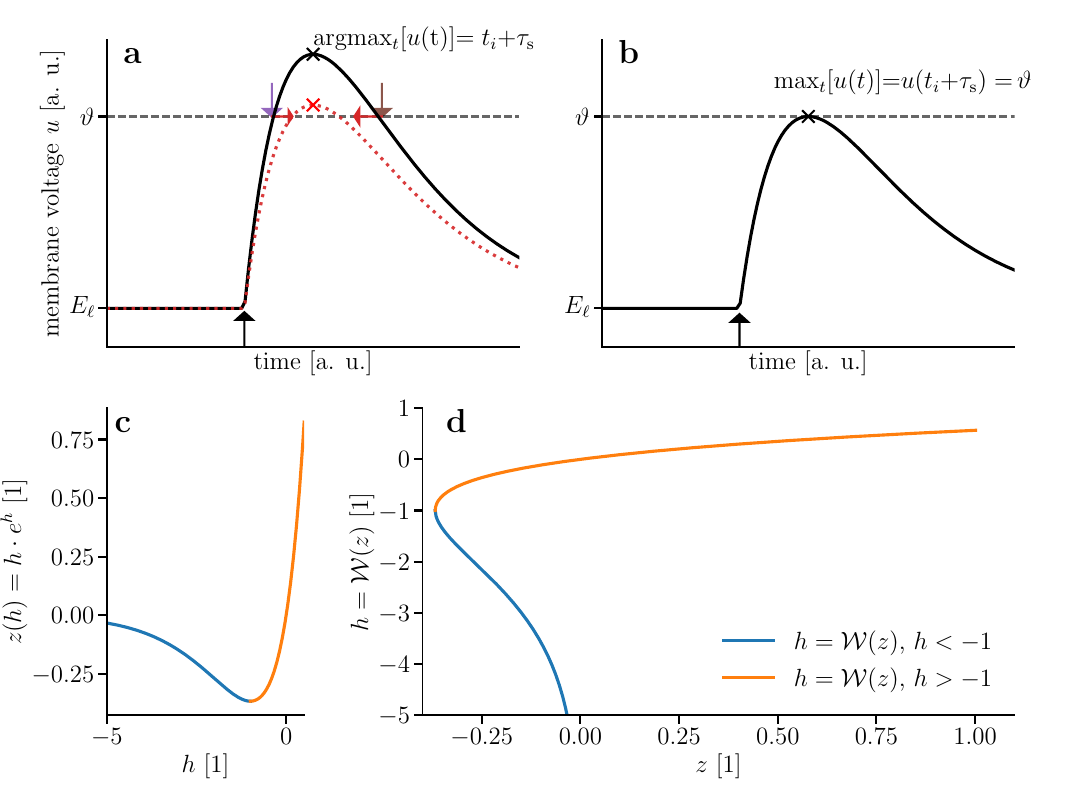}
        \caption[Choice of branch for case with $\taum=\taus$.]{
            \textbf{(a)} Membrane dynamics for one strong input spike at $t_i$ (upward arrow) with two threshold crossings due to leak pullback (earlier violet, later brown).
            The change induced by a reduction of the input weight is shown in red.
            \textbf{(b)} Edge case without crossing and exactly one time where $u(t)=\Vth$.
            \textbf{(c)} Defining relation for the Lambert W function $\mathcal{W}$, evidently not an injective map.
            \textbf{(d)} Distinguishing between $h\lessgtr-1$ allows to define the inverse function of (c), the Lambert W function $\mathcal{W}$.
        }	
        \label{fig:lambertw}
    \end{figure}
    \emph{Branch choice:}
    Given that a spike happens, there will be two threshold crossings: One from below at the actual spike time, and one from above when the voltage decays back to the leak potential (\cref{fig:lambertw}a,b).
    Correspondingly, the Lambert W function (\cref{fig:lambertw}c,d) has two real branches (in addition to infinite imaginary ones), and we need to choose the branch that returns the earlier solution.
    In case the voltage is only tangent to the threshold at its maximum, the Lambert W function only has one solution.
    
    For choosing the branch in the other cases we need to look at $h$ from the definition, i.e.
    \begin{equation}
            h = \frac y {a_1} = \frac b {a_1} - \frac T \taus.
    \end{equation}
    In a setting with only one strong enough input spike, the summations in $a_n$ and $b$ reduce to yield
    $h = (t_i - T) / \taus$.
    Because the maximum of the \gls{psp} for $\taum=\taus$ occurs at $t_i+\taus$, we know that the spike must occur at $T\leq t_i + \taus$ and therefore 
    \begin{equation}
        -1 \leq \frac{t_i - T}{\taus} = h.
    \end{equation}
    This corresponds to the branch cut of the Lambert W function meaning we must choose the branch with $h\geq -1$.
    For a general setting, if we know a spike exists, we expect $a_n$ and $b$ to be positive.
    In order to get the earlier threshold crossing, we need the branch that returns the larger $\mathcal{W}$ (\cref{fig:lambertw}d), that is where  $\mathcal{W}=h>-1$.

    \emph{Derivatives:}
    The derivatives for $t_i$ in the causal set $i\in C$ come down to
    \begin{align}
        \label{eq:appendix_equaltime_notinserted_dw}
        \frac{\partial T}{\partial w_i} & (\mathbf{w}, \mathbf{t}) \\
            = &\frac{\taus }{a_1}
                \exp \left( \frac{t_i}{\taus} \right)
                \left[ z\mathcal{W}'(z) + \left(\frac{t_i}{\taus} - \frac{b}{a_1}\right)
                        \left(1-z\mathcal{W}'(z) \right)\right]
            \,, \nonumber
        \\
        \label{eq:appendix_equaltime_notinserted_dt}
        \frac{\partial T}{\partial t_i} & (\mathbf{w}, \mathbf{t}) \\
            = &\frac{ w_i }{a_1} \exp\left(\frac{t_i}{\taus} \right) 
                \left[ 1 + \left(\frac{t_i}{\taus} - \frac{b}{a_1}\right)
                        \left(1-z\mathcal{W}'(z) \right)\right] \, .
            \nonumber
    \end{align}
    A crucial step is to reinsert the definition of the spike time where it is possible (cf.\ \cref{fig:stability}d).
    For this we need the derivative of the Lambert W function $z\mathcal{W}'(z) = \frach{\mathcal{W}(z)}{\mathcal{W}(z) + 1}$ that follows from differentiating its definition~\cref{eq:lambertw_def} with $h=\mathcal{W}(z)$ with respect to $z$.
    With this equation one can calculate the derivative of \cref{eq:appendix_equalTimeEquation} with respect to incoming weights and times as functions of presynaptic weights, input spike times and output spike time:
    \begin{align}
        \label{eq:appendix_equaltime_reinserted_dw}
        \frac{\partial T}{\partial w_i}  (\mathbf{w}, \mathbf{t}, T)
            &= - \frac{1}{a_1} \frac{1}{\mathcal{W}(z) + 1}
            \exp\left( \frac{t_i}{\taus} \right)
            \left(T - t_i\right)
            \,,
        \\
        \label{eq:appendix_equaltime_reinserted_dt}
        \frac{\partial T}{\partial t_i}  (\mathbf{w}, \mathbf{t}, T)
            &= - \frac{1}{a_1} \frac{1}{\mathcal{W}(z) + 1}
            \exp\left( \frac{t_i}{\taus} \right)
            \frac{w_i}{\taus}
            \left(T - t_i - \taus\right) \, .
    \end{align}
    These equations are equivalent to the \cref{eq:equaltime_dw_reinsert,eq:equaltime_dt} shown in the main text.

\paragraph
    [Learning rule for \texorpdfstring{$\taum=2\taus$}{time constants of ratio 2}]
    {Learning rule for $\boldsymbol{\taum=2\taus}$}
    Inserting the voltage (\crefp{eq:appendix_lif}) into the spike time (\crefp{eq:def_spike_time}) yields
    \begin{align}
        \gL\Vth = 
            &\exp\left(-\frac{T}{\taum}\right) \sum_{i\in C} w_i \exp\left(\frac{t_i}{\taum}\right)
            -\\&
            \exp\left(-\frac{T}{\taus}\right) \sum_{i\in C} w_i \exp\left(\frac{t_i}{\taus}\right)
            \,.\nonumber
    \end{align}
    Reordering and rewriting this in terms of $a_1$, $a_2$, and $\taus$ (with $\taum=2\taus$) we get
    \begin{equation}
        0 = 
            -a_1 \left[\exp\left(-\frac{T}{2\taus}\right)\right]^2
            + a_2 \exp\left(-\frac{T}{2\taus}\right)
            - \gL\Vth
            \,.
    \end{equation}
    This is written such that its quadratic nature becomes apparent, making it possible to solve for $\exp(-T/2\taus)$ and thus
    \begin{equation}
        \label{eq:appendix_doubleTimeEquation}
        \frac\T\taus
        = 
        2\ln \left[
            \frac{2a_1}{a_2 + \sqrt{a_2^2 - 4a_1\gL\Vth}}
        \right].
    \end{equation}
    
    \emph{Branch choice:}
    The quadratic equation has two solutions that correspond to the voltage crossing at spike time and relaxation towards the leak later; again, we want the earlier of the two solutions.
    It follows from the monotonicity of the logarithm that the earlier time is the one with the larger denominator.
    Due to an output spike requiring an excess of recent positively weighted input spikes, $a_n$ are positive, and the $+$ solution is the correct one.

    \emph{Derivatives:}
    \newcommand{\tmpx}{\sqrt{a_2^2 -4a_1\gL\Vth}}
    Using the definition $x=\tmpx$ for brevity, the derivatives of \cref{eq:appendix_doubleTimeEquation} are
    \begin{align}
        \label{eq:doubletime_deriv_notinserted}
        \frac{\partial T}{\partial w_i} & (\mathbf{w}, \mathbf{t})
        \\ 
        &=
        2\taus \left[ \frac{1}{a_1} + \frac{2\gL\Vth}{(a_2 + x)x} \right] 
        \exp \left( \frac{t_i}{\taus} \right)
            -\frach{2\taus}{x} \exp \left( \frac{t_i}{2\taus} \right) \, ,
            \nonumber
        \\
        \frac{\partial T}{\partial t_i} & (\mathbf{w}, \mathbf{t}) 
        \\&=
            2 w_i \left[ \frac{1}{a_1} + \frac{2\gL\Vth}{(a_2 + x)x} \right] 
            \exp \left( \frac{t_i}{\taus} \right)
            -\frach{w_i}{x} \exp \left( \frac{t_i}{2\taus} \right)
            . \nonumber
    \end{align}
    Again, inserting the output spike time yields
    \begin{align}
        \label{eq:doubletime_deriv_inserted}
        \frac{\partial T}{\partial w_i} & (\mathbf{w}, \mathbf{t}, T) 
        \\&=
            \frac{2\taus}{a_1} \left[ 1 + \frac{\gL\Vth}{x}
                \exp \left( \frac{T}{2\taus} \right)
            \right]
            \exp \left( \frac{t_i}{\taus} \right)
            -\frach{2\taus}{x} \exp \left( \frac{t_i}{2\taus} \right) \, ,
            \nonumber
        \\
        \frac{\partial T}{\partial t_i} & (\mathbf{w}, \mathbf{t}, T)
        \\&=
            \frac{2w_i}{a_1} \left[ 1 + \frac{\gL\Vth}{x}
                \exp \left( \frac{T}{2\taus} \right)
            \right] 
            \exp \left( \frac{t_i}{\taus} \right)
            -\frach{w_i}{x} \exp \left( \frac{t_i}{2\taus} \right)
            . \nonumber
    \end{align}

\paragraph{Error backpropagation in a layered network}
Our goal is to update the network's weights such that they minimize the loss function $L[\vect t^\br{N}, n^*]$.
For weights projecting into the label layer, updates are calculated via
\begin{equation}
    \label{eq:appendix_layer_top}
    \Delta w^\br{N}_{ni} 
    \propto - \frac{\partial L[\vect t^\br{N}, n^*]}{\partial w^\br{N}_{ni}} 
    = -
        \frac{\partial t^\br{N}_n}{\partial w^\br{N}_{ni}}
        \frac{\partial L[\vect t^\br{N}, n^*]}{\partial t^\br{N}_n} 
    \,.
\end{equation}
The weight updates of deeper layers can be calculated iteratively by application of the chain rule:
\begin{equation}
    \label{eq:appendix_layer_lower}
    \Delta w^{(l)}_{ki} \propto 
        - \frac{\partial L[\vect t^{(N)}, n^*]}{\partial w^{(l)}_{ki}} 
       = - \frac{\partial t^{(l)}_k}{\partial w^{(l)}_{ki}}  
       \delta^{(l)}_k
       \, ,
\end{equation}
where the second term is a propagated error that can be calculated recursively with a sum over the neurons in layer $(l+1$):
\begin{equation}
    \label{eq:appendix_layer_prop}
       \delta_k^\br{l}
       :=
       \frac{\partial L[\vect t^{(N)}, n^*]}{\partial t^{(l)}_{k}}
       =
            \sum_j \frac{\partial t^{(l+1)}_j}{\partial t^{(l)}_k}
            \delta^\br{l+1}_j
\, .
\end{equation}

In the following we treat the $\taum=\taus$ case but the calculations can be performed analogously for the other cases.
Rewriting \cref{eq:appendix_equaltime_reinserted_dt,eq:appendix_equaltime_reinserted_dw} in a  layer-wise setting, the derivatives of the spike time for a neuron $k$ in arbitrary layer $l$ are
\begin{align}
    \label{eq:appendix_equaltime_dw}
    \frac{\partial t^\br{l}_k}{\partial w^\br{l}_{ki}} &
    (\mathbf{w}, \mathbf{t}^{(l-1)}, \mathbf{t}^{(l)}) \\
     = - &\frac{1}{a_1} \exp \left( \dfrac{t^\br{l-1}_i}{\taus} \right) \frac{1}{\mathcal{W}(z) + 1} 
     \left(t^\br{l}_k - t^\br{l-1}_i\right) \, , \nonumber
    \\
    \label{eq:appendix_equaltime_dt}
	\frac{\partial t^\br{l}_k}{\partial t^\br{l-1}_i} &
        (\mathbf{w}, \mathbf{t}^{(l-1)}, \mathbf{t}^{(l)}) \\
     = - &\frac{1}{a_1} \exp \left( \frac{t^\br{l-1}_i}{\taus} \right)
        \frac{ 1 }{\mathcal{W}(z) + 1} 
    \frac{w^\br{l}_{ki}}{\taus}
    \left(t^\br{l}_k - t^\br{l-1}_i - \taus\right)
    \, . \nonumber
\end{align}
Inserting \cref{eq:appendix_layer_prop,eq:appendix_equaltime_dw,eq:appendix_equaltime_dt} into \cref{eq:appendix_layer_top,eq:appendix_layer_lower} yields a synaptic learning rule which implements exact error backpropagation on spike times.

This learning rule can be rewritten to resemble the standard error backpropagation algorithm for \glspl{ann}:
    \begin{align}
        \label{eq:appendix_compactTopError}
        \boldsymbol{\delta}^{(N)} &= \frac{\partial L}{\partial \boldsymbol{t}^{(N)}}\,, 
    \\
        \label{eq:appendix_compactErrorBP}
        \boldsymbol{\delta}^{(l-1)} &= 
            \left(\boldsymbol{\widehat{B}}^{(l)} -1\right) \odot
        \boldsymbol{\rho}^{(l-1)}\odot \left ( \boldsymbol{w}^{(l),T} \boldsymbol{\delta}^{(l)} \right )\,,
    \\
        \label{eq:appendix_compactWeightUpdate}
        \Delta \boldsymbol{w}^{(l)} &= -\eta \taus \left ( \boldsymbol{\delta}^{(l)} \boldsymbol{\rho}^{(l-1),T} \right ) \odot \boldsymbol{\widehat{B}}^{(l)}\,,
    \end{align}
where $\odot$ is the element-wise product, the $T$-superscript denotes the transpose of a matrix and $\boldsymbol{\delta}^\br{l-1}$ is a vector containing the  backpropagated errors of layer $\br{l-1}$.
The individual elements of the tensors above are given by
    \begin{align}
        \rho_i^{(l)} &= - \frac{1}{a_1} \exp\left(\frach{t_i^{(l)}}{\taus}\right) \frac{1}{\mathcal{W}(z) + 1} 
            \,, \\
        \widehat{B}_{ki}^{(l)} &= \frac{t_k^{(l)} - t_i^{(l-1)}}{\taus}
        \,.
    \end{align}

%% file: content/methods_emul.tex
The \gls{asic} is built around an analog neuromorphic core which emulates the dynamics of neurons and synapses.
All state variables, such as membrane potentials and synaptic currents, are physically represented in their respective circuits and evolve continuously in time.
Considering the natural time constants of such integrated analog circuits, this emulation takes place at 1000-fold accelerated time scales compared to the biological nervous system.
One \dls chip features 512 \gls{adex} neurons, which can be freely configured; these circuits can be restricted to \gls{lif} dynamics as required by our training framework \citep{aamir2018adex}.
Both the membrane and synaptic time constants were calibrated to \SI{6}{\micro\second}.

Each neuron circuit is connected to one of four synapse matrices on the chip, and integrates stimuli from its column of 256 \gls{cuba} synapses \citep{friedmann2016hybridlearning}.
Each synapse holds a \SI{6}{\bit} weight value; its sign is shared with all other synapses located on the same synaptic row.
The presented training scheme, however, allows weights to continuously transition between excitation and inhibition.
We therefore allocated pairs of synapse rows to convey the activity of single presynaptic partners, one configured for excitation, the other one for inhibition.

Synapses receive their inputs from an event routing module allowing to connect neurons within a chip as well as to inject stimuli from external sources.
Events emitted by the neuron circuits are annotated with a time stamp and then sent off-chip.
The neuromorphic \gls{asic} is accompanied by a \gls{fpga} to handle the communication with the host computer.
It also provides mechanisms for low-latency experiment control including the timed release of spike trains into the neuromorphic core.
The \gls{fpga} is furthermore used to record events and digitized membrane traces originating from the \gls{asic}.
\dls only permits recording one membrane trace at a time.
Each membrane voltage shown in \cref{fig:bss2}h therefore originates from a different repetition of the experiment.

The \gls{asic} is controlled by a layered software stack~\citep{muller2020extending} which exposes the necessary interfaces to a high-level user via Python bindings.
These were used in our framework that is described in the following.

%% file: content/methods_simul.tex
\paragraph{Simulation software}
Our experiments were performed using custom modules for the deep learning library \inlinecode{python}{PyTorch} \citep{torch_NEURIPS2019_9015}.
The network module implements layers of \gls{lif} neurons whose spike times are calculated according to \cref{eq:equalTimeEquation}. 
This method of determining the spike times of the neurons is fastest, but also memory-intensive.
An alternative implementation integrates the dynamical equations of the \gls{lif} neurons in a layer, which also yields the neuron spike times. 
Even though both approaches are technically equivalent, this method is slower and should only be employed if the computing resources are limited.

The activations passed between the layers during the forward pass  are the spike times.
The equations describing the weight updates for the network (\crefp{eq:appendix_compactWeightUpdate}) are realized in a custom backward-pass module for the network.

\paragraph{Training and regularization methods}

In order to train a given data set using our learning framework, the input data has to be translated into spike times first.
We do this by defining the times of the earliest and latest possible input spike $t_\text{early}$ and $t_\text{late}$ and mapping the range of input values linearly to the time interval $[t_\text{early}, t_\text{late}]$.
    
If the data set requires a bias to be solvable, our framework allows its addition.
These bias spikes essentially represent additional input spikes for a layer, which have the same spike time for any input.
The weights from the neurons to these ``bias sources" is learned in the same way as all the other synaptic weights.
For the Yin-Yang data set, the addition of a bias spike facilitated training.
For some samples, due to the low number of inputs, the relatively low activity that is received by the network is spread out over a long time interval.
The additional spike in the middle of the available interval decreases the maximum distance between input spikes for the hidden layer.
In contrast, the MNIST data set has a much higher input dimensionality and the spikes are more distributed over the input time interval.
Therefore, the activity provided to the hidden layer at any point in time is high even without additional bias.

Implementing our learning algorithm as custom PyTorch modules allows us to use the training architecture provided by the library. 
The simulations were performed using mini-batch training in combination with with the Adam optimizer~\citep{kingma2014adam} and learning rate scheduling (the parameters can be found in \cref{table:sim_params,table:hw_params}).

To assist learning we employ several regularization techniques.
The term $+\, \alpha \left[\exp\left(\fracl{t^\br{N}_{n^*}}{\beta\taus}\right) - 1 \right] $
with scaling parameters $\alpha, \beta\in\mathbb{R}^+$ , can be added to the loss in \cref{eq:loss}.
This regularizer further pushes the correct neuron towards earlier spike times. 

Gaussian noise on the input spike times can be used to combat overfitting.
This proved beneficial for the training of the MNIST data set.

Weight updates $\Delta w$ with absolute value larger than a given hyperparameter are set to zero to compensate divergence for vanishing denominator in~\cref{eq:appendix_compactWeightUpdate}.

As noted previously, the weight update equations are only defined for neurons that elicit a spike. 
To prevent fully quiescent networks we add a hyperparameter which controls how many neurons without an output spike are allowed.
If the portion of non-spiking neurons is above this threshold, we increase the input weights of the silent neurons.
In case of multiple layers where this applies, only the first such layer with insufficient spikes is boosted.
If neurons in a layer are too inactive multiple times in direct succession, the boost to the weights increases exponentially.

\paragraph{Training on hardware}
In principle our training framework can be used to train any neuromorphic hardware platform that (i) can receive a set of input spikes and yield the output spike times of all neurons in the emulated network and (ii) can update the weight configuration on the hardware according to the calculated weight updates.
In our framework the hardware replaces the computed forward-pass through the network.
For the calculation of the loss and the following backward pass, the hardware output spikes are treated  as if they had been produced by a forward pass in simulation.
The backward pass is identical to pure simulation.

As accessible value ranges of neuron parameters are typically determined by the hardware platform in use, a translation factor between the neuron parameters and weights in software and the parameters realized on hardware needs to be determined.
In our experiments with \dls the translation between hardware and software parameter domain was determined by matching of \gls{psp} shapes and spike times predicted by a software forward pass to the ones produced by the chip.

The implicit assumption of having only the first spike emitted by every neuron be relevant for downstream processing can effectively be ensured by using a long enough refractory period.
Since the only information-carrying signal that is not reset upon firing is the synaptic current, which is forgotten on the time scale of $\taus$, we found that, in practice, setting the refractory time $\tauref>\taus$ leads to most neurons eliciting only one spike before the classification of a given input pattern.

For training the Yin-Yang data set on \dls, having only five inputs proved insufficient due to the combination of limited weights and neuron variability.
We therefore multiplexed each logical input into five physical spike sources, totalling 25 inputs spikes per pattern.
Adding further copies of the inputs effectively increased the weights for each individual input.
This method has the added benefit of averaging out some of the effects of the fixed-pattern noise on the input circuits as multiple of them are employed for the same task.

\begin{table}
    \caption{
        Neuron, network and training parameters used to produce the results in \cref{fig:trainingChange,fig:sim_mnist}.
    }
    \centering
    \begin{threeparttable}
    \begin{tabular}{l|c|c}
        \textbf{Parameter name}  & \textbf{Yin-Yang} & \textbf{MNIST}\\
        \hline
        \textbf{Neuron parameters} &&\\
        $\gl$ & $1.0$ & $1.0$ \\
        $\Vleak$ & $0.0$ & $0.0$ \\ 
        $\Vth$ & $1.0$ & $1.0$ \\ 
        $\tau_\text{m}$ & $1.0$ & $1.0$ \\ 
        $\tau_\text{s}$ & $1.0$ & $1.0$ \\ 
        \hline
        \textbf{Network parameters} &&\\
        size input & $5$ & $784$ \\
        size hidden layer & $120$ & $350$ \\
        size output layer & $3$ & $10$ \\
        bias time\tnote{1} & $[0.9 \tau_\text{s}, 0.9 \tau_\text{s}]$ & no bias \\
        weight init mean\tnote{1} & $[1.5, 0.5]$ & $[0.05, 0.15]$ \\
        weight init stdev\tnote{1} & $[0.8, 0.8]$ & $[0.8, 0.8]$ \\
        $t_\text{early}$ & $0.15$ & $0.15$ \\
        $t_\text{late}$ & $2.0$ & $2.0$ \\
        \hline
        \textbf{Training parameters} &&\\
        training epochs & $300$ & $150$ \\
        batch size & $150$ & $80$ \\
        optimizer & Adam & Adam \\
        Adam parameter $\beta$ & $(0.9, 0.999)$ & $(0.9, 0.999)$ \\
        Adam parameter $\epsilon$ & $10^{-8}$ & $10^{-8}$ \\
        learning rate & $0.005$ & $0.005$ \\
        lr-scheduler & StepLR & StepLR \\
        lr-scheduler step size & $20$ & $15$ \\
        lr-scheduler $\gamma$ & $0.95$ & $0.9$ \\
        input noise $\sigma$ & no noise & $0.3$ \\
        max ratio missing spikes\tnote{1} & $[0.3, 0.0]$ & $[0.15, 0.05]$ \\
        max allowed $\Delta w$ & $0.2$ & $0.2$ \\
        weight bump value & $0.0005$ & $0.005$ \\
        $\alpha$ & $0.005$ & $0.005$ \\
        $\xi$\,\,\tnote{2} & $0.2$ & $0.2$ \\
        
    \end{tabular}
    \begin{tablenotes}
        \item[1] Parameter given layer wise [hidden layer, output layer].
        \item[2] $\xi$ implemented differently in code-base developed by the authors.
    \end{tablenotes}
    \end{threeparttable}
    \label{table:sim_params}
\end{table}

\begin{table}
    \caption{
        Network and training parameters for training on \dls used to produce the results in \cref{fig:bss2}. In contrast to \cref{table:sim_params}, the neuron parameters are not given here, as they are determined by the used chip.
    }
    \centering
    \begin{threeparttable}
    \begin{tabular}{l|c|c}
        \textbf{Parameter name}  & \textbf{Yin-Yang} & \textbf{16$\times$16 MNIST}\\
        \hline
        \textbf{Network parameters} &&\\
        size input & $25$ & $256$ \\
        size hidden layer & $120$ & $246$ \\
        size output layer & $3$ & $10$ \\
        bias time\tnote{1} & $[0.9 \tau_\text{s}, \text{no bias}]$ & no bias \\
        weight init mean\tnote{1} & $[0.1, 0.075]$ & $[0.01, 0.006]$ \\
        weight init stdev\tnote{1} & $[0.12, 0.15]$ & $[0.03, 0.1]$ \\
        $t_\text{early}$ & $0.15$ & $0.15$ \\
        $t_\text{late}$ & $2.0$ & $2.0$\tnote{3} \\
        \hline
        \textbf{Training parameters} &&\\
        training epochs & $400$ & $50$ \\
        batch size & $40$ & $50$ \\
        optimizer & Adam & Adam \\
        Adam parameter $\beta$ & $(0.9, 0.999)$ & $(0.9, 0.999)$ \\
        Adam parameter $\epsilon$ & $10^{-8}$ & $10^{-8}$ \\
        learning rate & $0.002$ & $0.003$ \\
        lr-scheduler & StepLR & StepLR \\
        lr-scheduler step size & $20$ & $10$ \\
        lr-scheduler $\gamma$ & $0.95$ & $0.9$ \\
        input noise $\sigma$ & no noise & $0.3$ \\
        max ratio missing spikes\tnote{1} & $[0.3, 0.05]$ & $[0.5, 0.5]$ \\
        max allowed $\Delta w$ & $0.2$ & $0.2$ \\
        weight bump value & $0.0005$ & $0.005$ \\
        $\alpha$ & $0.005$ & $0.005$ \\
        $\xi$\,\,\tnote{2} & $0.2$ & $0.2$ \\
        
    \end{tabular}
    \begin{tablenotes}
        \item[1] Parameter given layer wise [hidden layer, output layer].
        \item[2] $\xi$ implemented differently in code-base developed by the authors.
        \item[3] After translation of pixel values to spike times, inputs spikes with $t_\text{input} = t_\text{late}$ were not sent into the network.
    \end{tablenotes}
    \end{threeparttable}
    \label{table:hw_params}
\end{table}

%% file: content/ack.tex
We wish to thank Jakob Jordan and Nico Gürtler for valuable discussions, Sebastian Schmitt for his assistance with BrainScaleS-1, Vitali Karasenko, Philipp Spilger and Yannik Stradmann for taming physics, as well as Mike Davies and Intel for their ongoing support.
Some calculations were performed on UBELIX, the HPC cluster at the University of Bern.
Our work has greatly benefitted from access to the Fenix Infrastructure resources, which are partially funded from the European Union's Horizon 2020 research and innovation programme through the ICEI project under the grant agreement No. 800858.
Some simulations were performed on the bwForCluster NEMO, supported by the state of Baden-Württemberg through bwHPC and the German Research Foundation (DFG) through grant no INST 39/963-1 FUGG.
We gratefully acknowledge funding from the European Union under grant agreements 604102, 720270, 785907, 945539 (HBP) and the Manfred St{\"a}rk Foundation.

%% file: content/authcont.tex
JG, AB and MAP designed the conceptual and experimental approach.
JG derived the theory, implemented the algorithm, and performed the hardware experiments.
LK embedded the algorithm into a comprehensive training framework and performed the simulation experiments.
AB and OJB offered substantial software support.
SB, BC, JG and AFK provided low-level software for interfacing with the hardware.
JG, LK, DD, SB and MAP wrote the manuscript.

%% file: content/SI_bss1.tex
\begin{figure}[ht]
    \centering
    \includegraphics[width=0.48\textwidth]{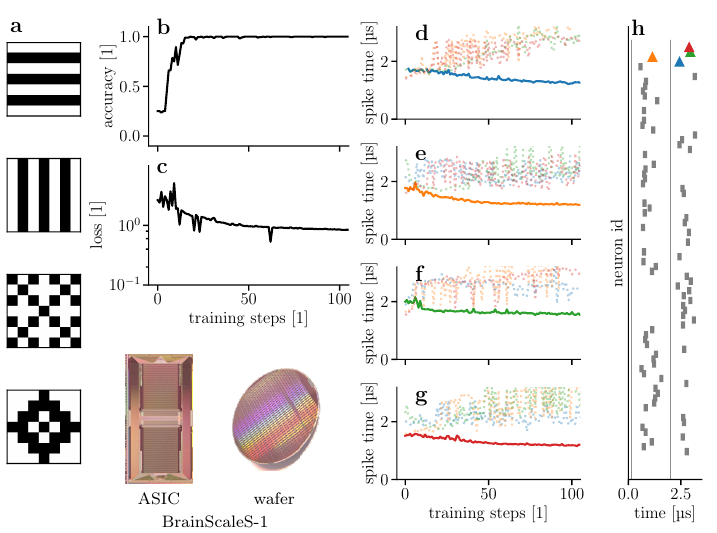}
	\caption{
	\textbf{Training a spiking network on the wafer-scale \bss system.}
            \textbf{(a)} Simple data set consisting of 4 classes with $7\times 7$ input pixels.
		    Accuracy \textbf{(b)} and  loss \textbf{(c)} during training of the four pattern data set.
		    \textbf{(d-g)} Evolution of the spike times in the label layer for the four different patterns.
		    In each, the neuron coding the correct class is shown with a solid line and in full color.
    		\textbf{(h)} Raster plot for the second pattern (e, correct class $\textcolor{mpl_c1}{\blacktriangle}$) after training.
	}	
	\label{fig:hardwareBSS1}
\end{figure}

To demonstrate the applicability of our approach to different neuromorphic substrates, we also tested it on the \bss system \citep{schemmel2010wafer}.
This version of BrainScaleS has a very similar architecture to \dls, but its component chips are interconnected through post-processing on their shared wafer (wafer-scale integration).
More importantly for our coding scheme and learning rules, its circuits emulate \gls{coba} instead of \gls{cuba} neurons.
Furthermore, due to the different fabrication technology and design choices \citep[in particular, the floating-gate parameter memory, see][]{srowig2007analog, schemmel2010wafer, Koke2017}, the parameter variability and spike time jitter are significantly higher than on \dls \citep{schmitt2017neuromorphic}.

The training procedure was analogous to the one used on \dls although using a different code base.
To accommodate the \gls{coba} synapse dynamics, we introduced global weight scale factors that modeled the distance between reversal and leak potentials and the total conductance, which were multiplied to the synaptic weights to achieve a \gls{cuba}.
This approximation could then be trained with our learning rules.
Despite this approximation and the considerable substrate variability, our framework was able to compensate well and classify the data set (\cref{fig:hardwareBSS1}) correctly after only few training steps.

%% file: content/SI_additional_results.tex
In addition to the simulation results collected in \cref{table:summary_of_results} we provide additional training results on the MNIST data set here (\cref{table:additional_results}).
We quantify the effect of noisy input spike times on generalization by comparing a noiseless
training run and a run with input noise, both using the hyperparameters shown in \cref{table:sim_params}.
Additionally, we train a network with a larger hidden layer as well as a deeper network with two hidden layers.
Finally, we illustrate the effect of the weight quantization on the training of the MNIST data set by using the same 6-bit quantization as on the \dls.

\begin{table}[h!]
    \caption{
    Additional simulation runs on the MNIST data set.
    The values given as the baseline are taken from \cref{table:summary_of_results}.
    With the noted exception of training length.
    Apart from the number of training epochs (see footnotes), the hyperparameters for simulations with the input resolution of $28\times28$ are the same as in \cref{table:sim_params} and the simulations for the input resolution of $16\times16$ used the hyperparameters given in \cref{table:hw_params}.
    }
    \centering
    \begin{threeparttable}
    \resizebox{\columnwidth}{!}{
    \begin{tabular}{l|c|c|c|c}
        \multirow{2}{*}{\textbf{simulation}} & \textbf{input}  & \textbf{hidden} & \multicolumn{2}{c}{\textbf{accuracy $[\si{\percent}]$}} \\
        & \textbf{resolution} & \textbf{neurons} & \textbf{test} & \textbf{train} \\[0.3em]
                \hline
        & & & \\[-0.7em]
        baseline & $28\times28$ & 350 & $97.1 \pm 0.1$ & $99.6 \pm 0.1$ \\
        without noise & $28\times28$ & 350 & $95.7 \pm 0.3$ & $99.7 \pm 0.1$ \\
        larger hidden layer & $28\times28$ & 800 & $97.3 \pm 0.1$ & $ 99.8 \pm 0.1$ \\
        two hidden layers\tnote{1} & $28\times28$ & 400-400 & $97.1 \pm 0.1$ & $ 99.5 \pm 0.1$ \\[0.6em]
        baseline\tnote{2} & $16\times16$ & 246 & $97.4 \pm 0.2$ & $ 99.2 \pm 0.1$ \\
        6-bit weight resolution\tnote{2} & $16\times16$ & 246 & $97.3 \pm 0.1$ & $ 99.1 \pm 0.1$ \\[0.6em]
    \end{tabular}
    }
    \begin{tablenotes}
        \item[1] \footnotesize This network was trained for 300 epochs.
        \item[2] \footnotesize This network was trained for 150 epochs.
    \end{tablenotes}
    \end{threeparttable}
    \label{table:additional_results}
\end{table}

%% file: content/SI_post_train_robustness.tex
\begin{figure}[ht]
    \centering
    \includegraphics[width=0.48\textwidth]{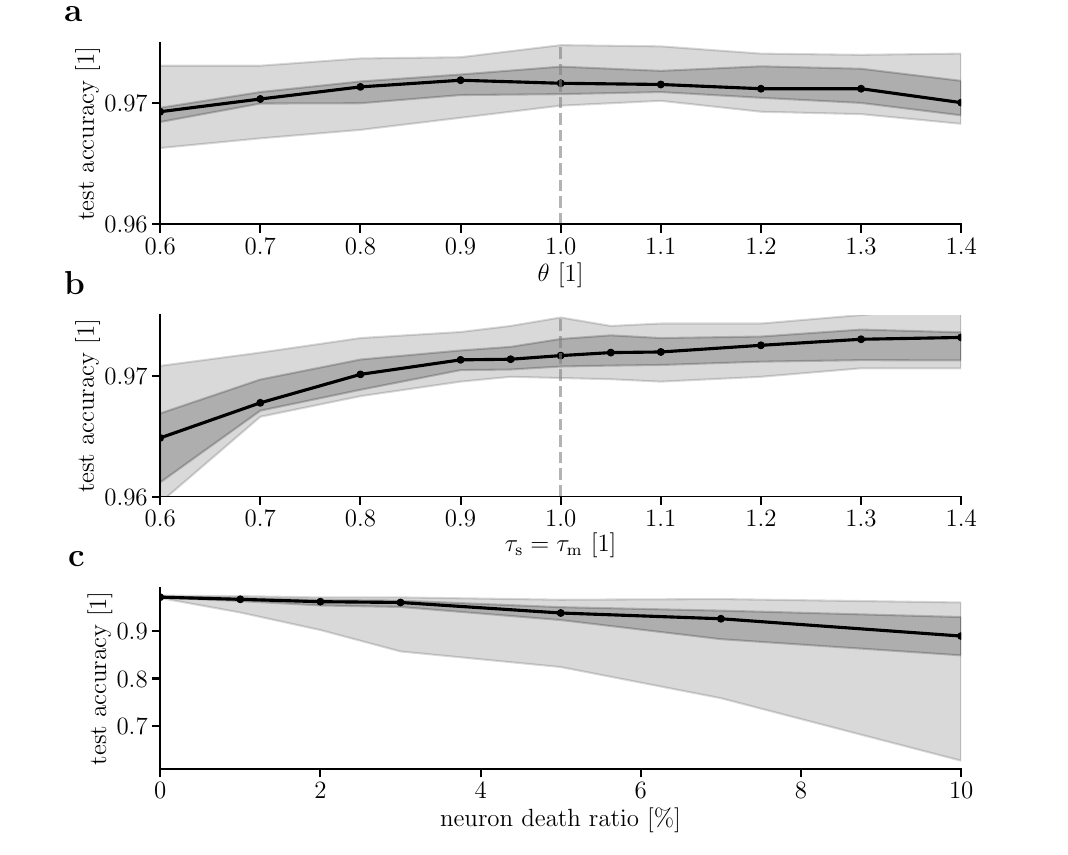}
	\caption{
		\textbf{Robustness to variations not present during training.}
	All panels show median (black), quartiles (dark gray), as well as the entire range between minimum and maximum (light gray) in the shaded regions.
	\textbf{(a)} Dependence of test accuracy for evaluation for 10 trained networks with shifted threshold value $\theta$.
	\textbf{(b)} Test accuracies for shifts in the neuron time constant $\taus$ and $\taum$.
	\textbf{(c)} Influence of random deletion of hidden neurons on test accuracies.
	For each neuron death ratio, 10 different random sets of hidden neurons were deleted.
	These ten deletion sets were applied to the same ten networks as in (a) and (b).
	}	
	\label{fig:postTrainRobustness}
\end{figure}

    We have already shown that our learning mechanism is able to cope well with noise and parameter variability which are present during training (\cref{fig:stability,fig:bss2}).
    In addition to these distortions which can be accounted for by the learning mechanism, it is interesting to measure the performance of the trained network under adverse effects that were not present during training.
    This is especially relevant for analog circuits where, for example, temperature changes can lead to shifts in the analog neuron parameters.
    We model this effect by training 10 networks on the MNIST data set using the ideal parameters of $\Vth = 1$ and $\taus = \taum = 1$ for the neuron threshold and time constant and then evaluating their performance on the test data set for shifted values of the threshold and time constant (\cref{fig:postTrainRobustness} a, b).
    These simulations show that the trained networks cope well, even if the relative shifts to the parameters are much larger than what can be typically expected due to temperature changes on \dls.

    Furthermore, we consider a scenario which is less likely on neuromorphic platforms, but may be more relevant in biological networks.
    In biology, neural networks have to be robust against the death of neuron cells within the network.
    For each of the 10 fully trained networks we delete a percentage of its hidden population and evaluate the performance on the test set.
    As the consequences of this procedure strongly depend on exact choice of the deleted neurons, we repeat each deletion scenario for each network 10 times with different random seeds.
    \Cref{fig:postTrainRobustness}c shows that networks trained with our learning mechanism exhibit stability towards sudden neuron death after training and even for \SI{5}{\percent} neuron death the bound of the second quartile is still at \SI{92.3}{\percent} accuracy.
    Note also that if plasticity is ongoing, the network can learn to recover much of its performance after apoptosis.

%% file: content/SI_LRsimplified.tex
\begin{figure}[ht]
    \centering
    \includegraphics[width=0.48\textwidth]{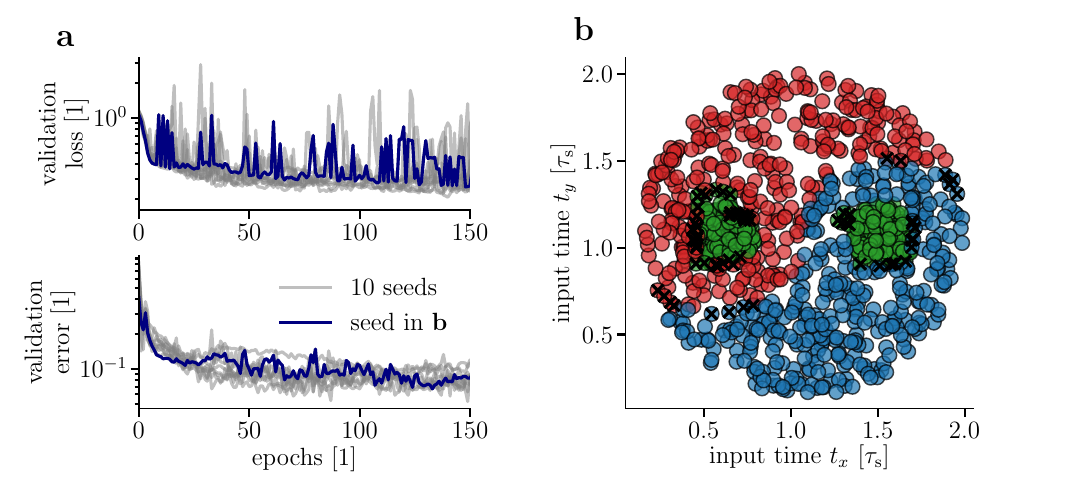}
	\caption{
        \textbf{Training on the Yin-Yang data set with a simplified learning rule.}
        We approximated the learning rule to have less complex updates (\crefpPlTwo{eq:si_simplifiedLR_dw}{eq:si_simplifiedLR_dt}).
        \textbf{(a)} shows the training process of 150 epochs. We reach a test accuracy of \SI{91.7\pm1.4}{\percent} and training accuracy \SI{91.7\pm1.2}{\percent} averaged over 10 seeds.
        \textbf{(b)} shows the classification as in~\cref{fig:trainingChange} after training for the highlighted training in (a).
	}\label{fig:SI_simplifiedLR_YY}
\end{figure}

        The learning rule for $\taum=\taus$ described in the main paper and derived in the \methods\ is computationally rather demanding:
        it needs multiple evaluations of the exponential function as well as an evaluation of the Lambert W function $\mathcal{W}$, for which no closed form exists.
        As the computational complexity of plasticity mechanisms on many neuromorphic chips is limited, we investigate the possibility of approximating the derivatives \cref{eq:equaltime_dw_reinsert,eq:equaltime_dt} by replacing the exponential functions as well as $\mathcal{W}$ by a constant 
    
        $\lambda$\footnote{This effectively leads to $\rho$ being a constant in \cref{eq:appendix_compactErrorBP,eq:appendix_compactWeightUpdate}.}:

    \begin{align}
        \label{eq:si_simplifiedLR_dw}
        \frac{\partial t_k}{\partial w_{ki}}
         &= - \lambda
         \left(t_k - t_i\right) \, ,
         \\
        \label{eq:si_simplifiedLR_dt}
        \frac{\partial t_k}{\partial t_i}\,\,
         & = - \lambda
        \frac{w_{ki}}{\taus}
        \left(t_k - t_i - \taus\right)
        \, .
    \end{align}

        The approximated version consists only of simple differences and multiplications making this learning rule more amenable for on-chip implementations.

        To examine the approximated learning rule in the standard setup with $\taum=\taus$  we chose $\lambda=0.0192$ by evaluating $\frac{1}{a_1} \frac{1}{\mathcal{W}(z) + 1}$ for a few inputs into the hidden layer.
        Using this extreme simplification we trained a network to classify the Yin-Yang data set (\cref{fig:SI_simplifiedLR_YY}).
        While the network learned the task correctly and achieved a test accuracy of \SI{91.7\pm1.4}{\percent}, this represents a small but noticeable
        drop in accuracy compared to the full learning rule (\cref{table:summary_of_results}).
        We also observed that these simplified rules led to more instability for longer training periods (not shown here).
        Nonetheless, these promising results give us confidence that that a more careful choice of the constant or a replacement with a simple, but non-constant term can alleviate these problems while retaining a simple form of the learning rule.
        
        Note, in particular, that \cref{eq:si_simplifiedLR_dt} explicitly contains the term $t_i + \taus$.
        This term represents the maximum of a \gls{psp} and changes sign when the output spike at $T$ happens before versus after the maximum.
        This simple difference captures the major non-monotonic relationship in the time derivative.
        As the maximum of the \gls{psp} is given by a closed form solution $t_\mathrm{max} = t_i + \frac{\taum\taus}{\taus-\taum}  \log \frac\taus\taum$ for arbitrary combinations of $\taus$ and $\taum$, it seems natural to investigate a slightly altered learning rule for different time constants.

%% file: content/SI_energy.tex
    \Cref{table:energy_reference} in the main manuscript compares the energy consumption and classification speed of our model on \dls with other neuromorphic platforms and an ANN on a GPU.
    This section details how the power and classification speed measurements were performed, as well as
    their implications for the scalability of and potential improvements to our setup.
    Additionally, we present our measurement technique for the GPU reference.

\subsubsection{\dls}\label[supplementary]{subsec:SI_energy_bss}
\paragraph{Power breakdown}
    
    The full BrainScaleS-2 chip consumed a total of \SI{175}{\milli\watt} measured during runtime with the INA219 chip \cite{INA219:datasheet}.
    This overall figure encompasses the chip's high-speed communication links (approx. \SI{60}{\milli\watt}), the digital periphery as well as its clocking infrastructure (approx. \SI{80}{\milli\watt}), and the biasing of analog circuits (approx. \SI{35}{\milli\watt}).
    Importantly, we could not observe a significant change in power consumption between the network during inference and an emulation of an inactive network.
    This implies that the cost of event transport and synaptic processing is negligible on the reported scales and that the overall figure would not be impacted by increased activity levels.
    As inactive synapses mostly contribute to the overall power consumption through negligible leakage currents, the power consumption would not be impacted by an increase of the neuron circuit's fan-in that would allow the training on larger input spaces.

\paragraph{Execution time breakdown}
    
    We define the round-trip time for an on-chip inference run as starting before the forward pass through the network in our PyTorch implementation and ending when all classification results produced by the chip are available in PyTorch.
    For the classification of the full MNIST test data set on \dls we measured a round-trip time of \SI{0.937}{\second}.
    
\begin{figure}[ht]
    \centering
    \includegraphics[width=0.48\textwidth]{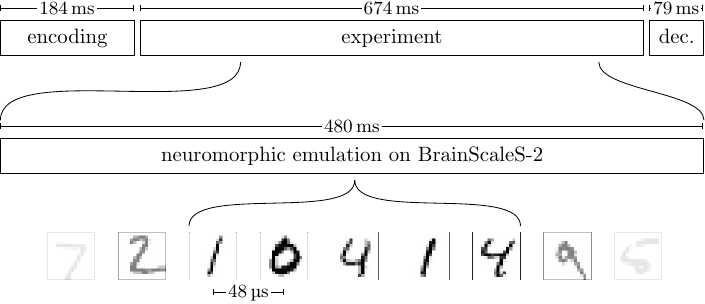}
	\caption{
	\textbf{Breakdown of the execution time for inference on the MNIST test set.}
	The total time of  about one second consists of an encoding, an experiment and a decoding phase.
	The encoding phase includes the translation of PyTorch tensors into spike trains and the encoding of the spike trains into instructions for the neuromorphic chip.
	In the experiment phase the instructions are sent from the host to the chip, the emulation is performed and the results are read out from the chip and communicated back to the host.
	In the final decoding phase the emulations results are converted back to PyTorch tensors.
	}	
    \label{fig:SI_energy_timingBSS}
\end{figure}

    Due to this conservative definition of the round-trip time, our measurement includes a significant amount of time on the host (for data pre- and post-processing) and for communication between host and the neuromorphic system.
    \cref{fig:SI_energy_timingBSS} shows a detailed breakdown of the execution time.
    We see that once the data arrives on the chip, it takes \SI{480}{\milli \second} to process the $10\,000$ images of the test set.
    This results in a classification every \SI{48}{\micro\second} or equivalently a classification rate of $20\,800$ images per second.
    
    Considering the relevant hardware time constants of $\taus \approx \taum \approx...\SI{6}{\micro\second}$  and the typical time to solution of around $1\,\taus$ to $1.5\,\taus$, a classification duration per sample of \SI{48}{\micro\second} seems surprisingly long. 
    This is owed to the sequential presentation of data samples to the network, for which we need to ensure that all residual activity  -- membrane voltages as well as synaptic currents -- from the last sample has fully decayed before the next sample is presented.
    Currently, this is achieved by simply waiting between samples, but \citet{cramer2020training} present an alternative:
    The \gls{ppu} is able to trigger a reset of all membrane voltages and synaptic currents on the chip.
    Using this mechanism, \citet{cramer2020training} shorten the classification time per image to \SI{11.8}{\micro\second}.
    The usage of artificial resets would also be a viable optimization for our model.
    It would require the previously switched off \gls{ppu} to be activated and would therefore slightly increase the power consumption (by approximately \SI{20}{\milli \watt}).
    This increase in power would however be more than compensated  by the approximately quadrupled sample throughput.

\subsubsection{GPU}\label[supplementary]{subsec:SI_energy_vN}

For comparison to conventional computing hardware we add power and classification speed measurements on a Nvidia Tesla P100 GPU to \cref{table:energy_reference}.
For the measurements on the GPU we use the convolutional neural network given as standard example in the PyTorch repository (\url{https://github.com/pytorch/examples/blob/507493d7b5fab51d55af88c5df9eadceb144fb67/mnist/main.py}).

\begin{figure}[ht]
    \centering
    \includegraphics[width=0.48\textwidth]{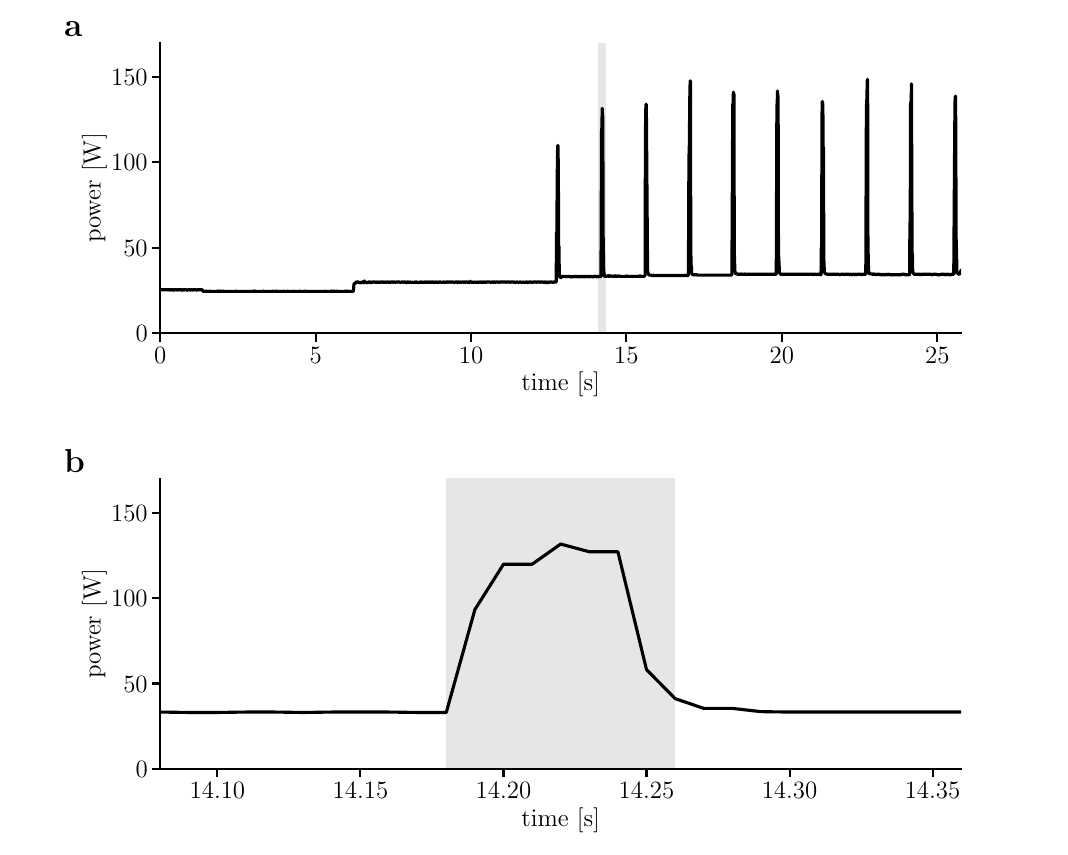}
	\caption{
	\textbf{Power consumption of Nvidia Tesla P100 GPU during classification of MNIST test data.}
	\textbf{(a)} Power consumption   of a standard PyTorch network for MNIST classification while running inference on the test data set for $10$ times.
	\textbf{(b)} Zoom on a peak in the power consumption. The shaded area corresponds to the time during which the GPU is actively used (measured from within Python). 
	}	
	\label{fig:gpu_power}
\end{figure}

The power measurements are performed using the tool \inlinecode{python}{nvidia-smi} which is runnning in the background while in the foreground we run a PyTorch program which repeats the classification of the MNIST test data set for $10$ times.
\Cref{fig:gpu_power} shows the power consumption over the full program runtime.
We see that the GPU is only active for 10 short periods, the duration of which match the measured times during which the PyTorch program uses the GPU (\cref{fig:gpu_power} b).
The power consumption is calculated as an average over these intervals, resulting in \SI{106.5}{\watt}.

The speed measurements were performed using time measurements in Python and averaged over the 10 classifications, resulting in a classification time per image of \SI{8}{\micro\second}.
This amounts to an energy-per-classification value of \SI{852}{\micro\joule}.

As an additional reference we attempted to determine the power consumption and classification speed for a fully connected network with a hidden layer of 246 neurons (same size as the hidden layer on \dls) on GPU.
However, due to the fact that the classification was a factor of 20 to 25 faster than for the CNN, 
we were not able to measure the power in a fine enough resolution with \inlinecode{python}{nvidia-smi} to yield reliable values.
To estimate a lower-bound for the energy per classification in this case, we can take the power consumption of the GPU in the phases where it was not actively used in the CNN measurement (i.e. power values between the peaks in \cref{fig:gpu_power}a) which is approximately \SI{34}{\watt}.
This ``idle" power consumption for the CNN case seemed to approximately match the averaged power drain for the fully connected network.
This amounts to a lower-bound estimate of the energy-per-classification value on the order of \SI{10}{\micro\joule}.

%% file: content/SI_table.tex
\begin{table*}[!htbp]
    \caption{
    Extension of literature review for pattern recognition models on neuromorphic back-ends, including results which do not detail certain measurements.
    }
    \centering
    \begin{threeparttable}
    \resizebox{\textwidth}{!}{
    \begin{tabular}{lllccccc}
        \multirow{2}{*}{\textbf{platform}} & \multirow{2}{*}{\textbf{type}}  & \multirow{2}{*}{\textbf{coding}} & \textbf{network} & \textbf{energy per} & \textbf{classifications} & \textbf{test} & \multirow{2}{*}{\textbf{reference}}\\
        &&& \textbf{size/structure} & \textbf{classification} & \textbf{per second} & \textbf{accuracy} & \\[0.3em]
        \hline
        &&&&&\\[-0.7em]
        SpiNNaker & digital & rate & 764-600-500-10 & \SI{3.3}{\milli\joule} & $91$  & \SI{95.0}{\percent} & \cite{stromatias2015scalable}, 2015\\[0.3em]
        True North\tnote{1} & digital & rate & CNN & \SI{0.27}{\micro\joule} & $1000$ & \SI{92.7}{\percent} & \cite{esser2015backpropagation}, 2015 \\[0.3em]
        True North\tnote{1} & digital & rate & CNN & \SI{108}{\micro\joule} & $1000$ & \SI{99.4}{\percent} & \cite{esser2015backpropagation}, 2015\\[0.3em]
        FPGA (nLIF neurons)\tnote{2} & digital & temporal & 784-600-10 & - & - & \SI{96.8}{\percent} & \cite{mostafa2017fpga}, 2017\\[0.3em]
        unnamed (Intel)\tnote{3} & digital & temporal & 236-20 & \SI{17.1}{\micro\joule} & $6250$ & \SI{89.0}{\percent} & \cite{chen20184096}, 2018 \\[0.3em]
        unnamed (Intel)\tnote{4}			& digital	& temporal	& 784-1024-512-10 & \SI{112.4}{\micro\joule} & - &  \SI{98.2}{\percent}	& \cite{chen20184096}, 2018 \\[0.3em]
        unnamed (Intel)\tnote{4} & digital	& temporal	& 784-1024-512-10 & \SI{1.7}{\micro\joule}	& - & \SI{97.9}{\percent}	& \cite{chen20184096}, 2018 \\[0.3em]
        Loihi \tnote{5} & digital	& temporal	& 1920-10 & - & - & \SI{96.4}{\percent}	& \cite{lin2018programming}, 2018 \\[0.3em]
        SPOON \tnote{6} & digital	& temporal	& CNN & \SI{0.3}{\micro\joule} & 8547 & \SI{97.5}{\percent}	& \cite{frenkel202028}, 2020 \\[0.3em]
        \dls & mixed & temporal & 256-246-10 & \SI{8.4}{\micro\joule} & $20\,800$ & \SI{96.9}{\percent} & this work \\[1.0em]
    \end{tabular}
    }
    \begin{tablenotes}
        \item[1] In \cite{esser2015backpropagation} it is stated that "The instrumentation available measures active power for the network in operation and leakage power for the entire chip, which consists of 4096 cores. We report energy numbers as active power plus the fraction of leakage power for the cores in use.". For the first result 5 cores were used, while the second result requires 1920 cores.
        \item[2] No energy or speed measurements reported.
        \item[3] Images preprocessed with $4$ $5\times 5$ Gabor filters and $3\times 3$ pooling.
        \item[4] No speed measurements reported.
        \item[5] No energy or speed measurements reported.
        Images were preprocessed with an algorithm described as ''using scan-line encoders".
        \item[6] Reported energy values are pre-silicon simulations.
    \end{tablenotes}
    \end{threeparttable}
    \label[supplementarytable]{table:appendix_fullLiterature}
\end{table*}